\documentclass{article} 
\usepackage{iclr2026_conference,times}

\usepackage{amsmath,amsfonts,bm}

\def\eqref#1{equation~\ref{#1}}

\def\1{\bm{1}}

\DeclareMathAlphabet{\mathsfit}{\encodingdefault}{\sfdefault}{m}{sl}
\SetMathAlphabet{\mathsfit}{bold}{\encodingdefault}{\sfdefault}{bx}{n}

\usepackage{hyperref}
\usepackage{url}

\usepackage{graphicx}
\usepackage{wrapfig}
\usepackage{multirow}
\usepackage{tabularx}
\usepackage{booktabs}
\usepackage{textcomp}
\usepackage{xcolor}
\usepackage{float}
 
\usepackage{subcaption}   
\usepackage{colortbl}
\usepackage{amsmath}
\usepackage{hhline}

\usepackage{algorithm}
\usepackage{algpseudocode}
\usepackage{varwidth}

\title{ETC: training-free diffusion models acceleration with Error-aware Trend Consistency}

\author{
    Jiajian Xie$^{1*}$ \quad Hubery Yin$^2$ \quad Chen Li$^2$ \quad Zhou Zhao$^1$ \quad Shengyu Zhang$^{1\dag}$ \\
    $^1$Zhejiang University \quad 
    $^2$WeChat Vision, Tencent Inc. \\
    \{xiejiajian, zhaozhou, sy\_zhang\}@zju.edu.cn, \{hubery, chaselli\}@tencent.com
}

\iclrfinalcopy 
\begin{document}

\maketitle
\footnotetext{$*$This work was done when Jiajian Xie was an intern at WeChat.\quad $\dagger$ Corresponding author. }

\begin{abstract}
Diffusion models have achieved remarkable generative quality but remain bottlenecked by costly iterative sampling. Recent training-free methods accelerate diffusion process by reusing model outputs. However, these methods ignore denoising trends and lack error control for model-specific tolerance, leading to trajectory deviations under multi-step reuse and exacerbating inconsistencies in the generated results. To address these issues, we introduce Error-aware Trend Consistency (ETC), a framework that (1) introduces a consistent trend predictor that leverages the smooth continuity of diffusion trajectories, projecting historical denoising patterns into stable future directions and progressively distributing them across multiple approximation steps to achieve acceleration without deviating; (2) proposes a model-specific error tolerance search mechanism that derives corrective thresholds by identifying transition points from volatile semantic planning to stable quality refinement. Experiments show that ETC achieves a 2.65× acceleration over FLUX with negligible (-0.074 SSIM score) degradation of consistency.
\end{abstract} 


\section{Introduction}
Diffusion models~\citep{sohl2015deep,song2019generative,ho2020denoising} have demonstrated remarkable generative capabilities across diverse domains including images, videos and audio. However, their superior generative performance typically requires larger model architectures and multi-step denoising processes, resulting in substantial computational overhead and inference latency. Training-based methods~\citep{salimans2022progressive,luo2023latent} accelerate diffusion by learning a few-step model from the denoising process of a multi-step model. However, this paradigm requires extensive training and there often remains a discrepancy between the predictive distributions of the original and the few-step model~\citep{stanton2021does}, weakening generalization capabilities.

\begin{figure}[t]
	\centering
	\begin{subfigure}{0.452\textwidth}
		\centering
		\includegraphics[width=\linewidth]{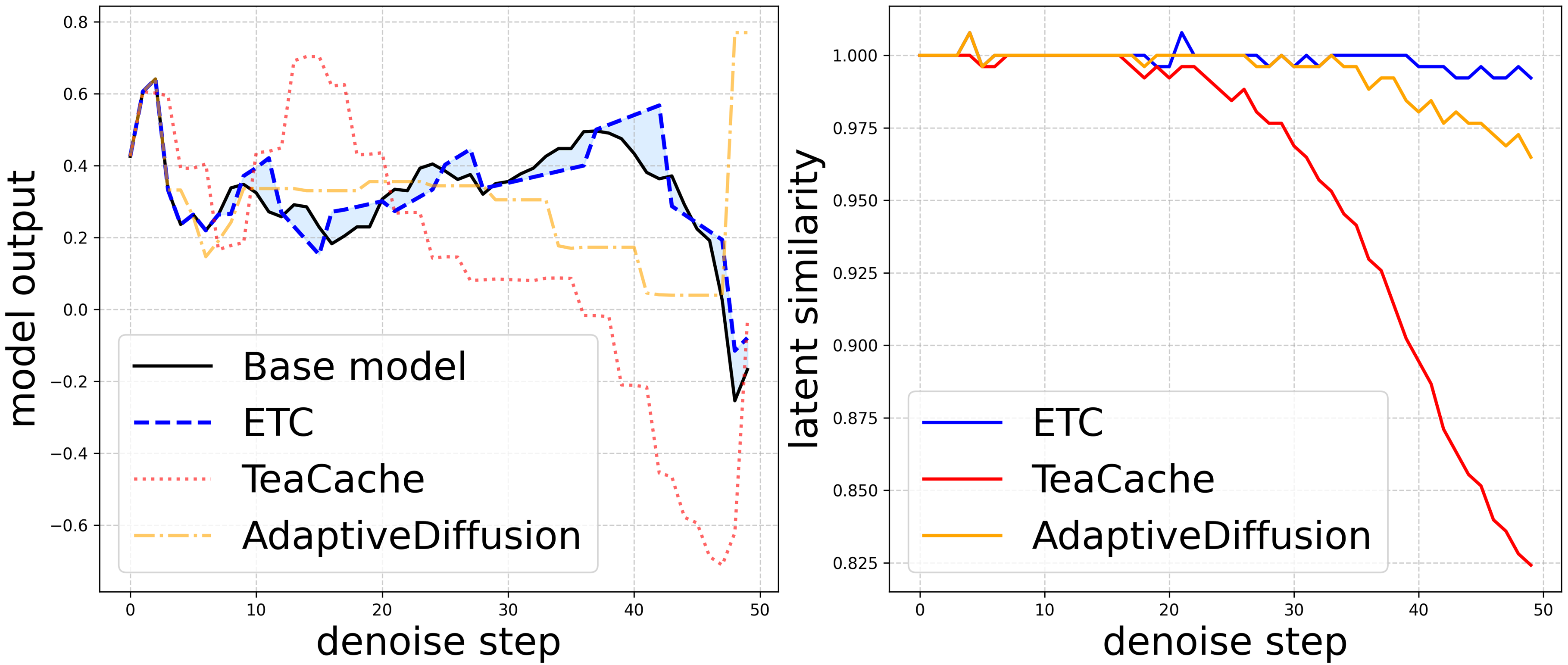}
		\caption{Comparison of denoising trajectories}
		\label{fig:consis}
	\end{subfigure}
	\begin{subfigure}{0.54\textwidth}
		\centering
		\includegraphics[width=\linewidth]{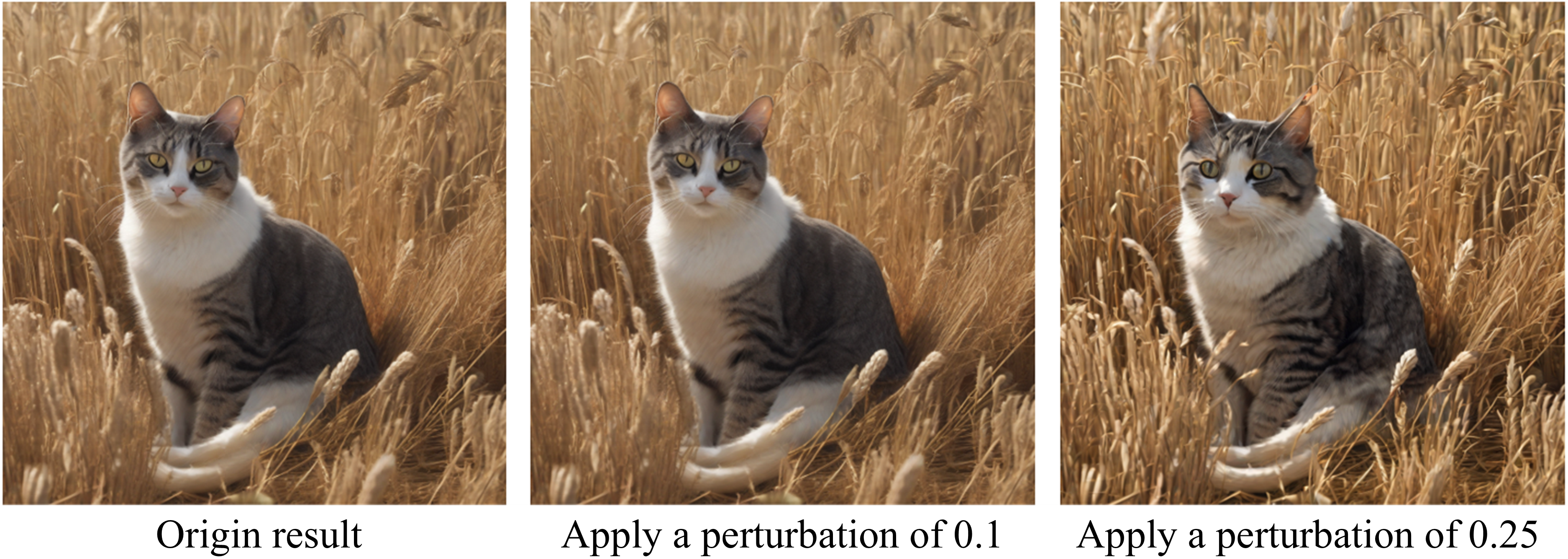}
		\caption{Effect of different levels of denoising errors}
		\label{fig:error-correct}
	\end{subfigure}
	\caption{Visualization of trajectory deviation and denoise error tolerance. Subfigure (a) shows that existing methods fail to follow the original denoising trajectory and reduce latent similarity. Subfigure (b) shows the model maintains consistent results to a certain degree of denoising errors.}
	\vspace{-0.3cm}
	\label{fig:intro}
\end{figure}

In contrast, training-free methods~\citep{chen2406delta,ye2024training} accelerate diffusion without model retraining or performance degradation by leveraging feature similarity between adjacent timesteps, broadly categorized into inner-level and step-wise feature reuse mechanisms. Inner-level~\citep{ma2024learning,liu2024faster} methods accelerate inference by reusing internal layer features within each denoising iteration but require architecture-specific designs. In contrast, step-wise~\citep{ye2024training,liu2025timestep} methods evaluate model output robustness to determine multi-step reuse and reduce the total number of model inferences, achieving better generalized acceleration compared to inner-level alternatives. However, step-wise methods overlook denoising trends between adjacent timesteps, leading to inconsistent generation results. While recent approaches~\citep{chen2406delta,liu2025timestep} attempt to address this by employing residuals between model outputs for trend prediction, short-term output fluctuations often deviate from the long-term denoising trajectory, thereby exacerbating trend inconsistency caused by the multi-step use of fluctuating residuals as shown in Figure~\ref{fig:consis}. Consequently, achieving \textbf{multi-step approximations with consistent trajectories} remains an unresolved challenge. Moreover, existing methods rely on manually-defined fixed thresholds for approximation error assessment, neglecting model-specific error tolerances shown in Figure~\ref{fig:error-correct}. Consequently, fixed threshold strategy may fails to address scenarios where accumulated errors cause irreversible trajectory deviations beyond the model's corrective capacity, resulting in \textbf{inadequate error control} and amplified generation inconsistency. To summarize, step-wise feature reuse research demonstrates significant untapped acceleration potential but is challenged with trajectory deviations under multi-step approximation due to the lack of accurate capture of denoising trends, and inadequate error control due to insufficient exploration of model error tolerance limits.

To address the above problems, we propose ETC, a framework that achieves step-wise diffusion acceleration through trajectory consistency and model-specific error control. For the multi-step trajectory consistency problem, we aim to derive future directional flows from the global stabilization properties inherent in the denoising process. Specifically, we compute weighted projections of cross-step changes in model outputs, attenuating fluctuations caused by approximation errors while amplifying directional components of long-term dynamics. To further enhance acceleration, we design an adaptive expansion strategy that dynamically extends or contracts the approximation window based on whether deviations between projected trends and periodic model inferences remain within the model’s corrective capacity. Moreover, we employ a progressive distribution paradigm that proportionally allocates estimated direction flows across each approximation step, thereby achieving aggressive acceleration while preventing divergence from the denoising trajectory. As for the inadequate error control problem, we treat different models' error tolerance as emergent properties within denoising dynamics. Observing that the semantic planning phase in denoising process exhibit high dynamic variance~\citep{liu2024faster}, we quantify the perceptual influence of deviation perturbations on generation quality and derive critical transition points from volatile semantic planning to smooth quality refinement phases that reflect model limits for error correction.

In summary, the main contributions of out work are as follows: (1) We introduce a consistent trend predictor that projects stabilizing denoising trends into future flows and progressively distributes them across approximation steps to ensure trajectory consistency. (2) We propose a model-specific error tolerance search method that quantifies the impact of accumulated errors and locates the transition point between volatile semantic planning and stable quality refinement to reflect the error tolerance threshold. (3) Experiments show that ETC outperforms other state-of-the-art baselines in terms of generation consistency and speed across images, videos and audio synthesis.

\vspace{-0.1cm}
\section{RELATED WORK}
\vspace{-0.1cm}
\noindent\textbf{Diffusion model}\quad Diffusion models have achieved great success in generative tasks. Early diffusion models~\citep{ho2020denoising,song2020denoising} implemented iterative denoising processes directly on the original data modality, which imposed computational overhead due to high-dimensional operations. To address computational efficiency constraints, latent diffusion models~\citep{rombach2022high,blattmann2023align} compress the original modality into lower-dimensional representations before applying the diffusion process. Initially developed with the U-Net~\citep{ronneberger2015u}, latent diffusion models have demonstrated impressive performance. However, U-Net-based design encountered scalability constraints that limited larger model training and practical deployment. Diffusion Transformers (DiT)~\citep{peebles2023scalable} address this limitation by leveraging transformer~\citep{vaswani2017attention} for enhanced scalability and achieve state of-the-art performance across diverse domains~\citep{flux2024,wan2025wan,hung2024tangoflux}. Despite the high quality achieved by diffusion models, the multi-step denoising design slows down the inference process.

\noindent\textbf{Diffusion model acceleration}\quad Diffusion model acceleration can be categorized into training-based and training-free approaches. Training-based methods aim to learn a few-step model from the multi-step denoising process. Progressive distillation~\citep{salimans2022progressive} progressively matches noise predictions between teacher and student models. Latent Consistency Model (LCM)~\citep{luo2023latent} achieves single-step sampling by imposing self-consistency constraints. However, these methods require time-consuming training and often suffer performance degradation due to the gap between the predictive distributions of the few-step and original models~\citep{stanton2021does}.

In contrast, recent training-free approaches leverage feature reuse between adjacent time steps for acceleration and can be divided into inner-level and step-wise feature reuse. Inner-level reuse accelerates diffusion by reusing features within different layers of the model. DeepCache~\citep{ma2024deepcache} caches high-level features in the U-Net while dynamically updating only shallow features in subsequent steps. T-Gate~\citep{liu2024faster} caches DiT cross-attention outputs after convergence and keeps them fixed for remaining steps. However, these methods are model-specifically designed and lack generalizability. Step-wise methods exploit cross-temporal output similarity for general acceleration. AdaptiveDiffusion~\citep{ye2024training} detects redundancy across denoising steps using a third-order differential estimator and reuses historical model outputs. However, it neglects the denoising trend and leads to gradual trajectory deviations. TeaCache~\citep{liu2025timestep} minimizes trajectory deviation by using residuals between adjacent steps as approximate trends, but it may amplify error accumulation due to short-term denoising fluctuations. SADA~\citep{jiang2025sada} enhances trend estimation by combining three-step historical results, yet its acceleration potential is limited by a maximum four-step jump. Although step-wise methods show acceleration potential, maintaining trend consistency in multi-step approximations remains a challenge. Furthermore, existing step-wise methods rely on manually predefined thresholds to determine the reusability, resulting in inconsistent performance across different models due to the lack of model-specific thresholds.

In this paper, we focus on maintaining the consistency of denoised trajectories during step-wise acceleration. We combine all historical trend patterns to obtain more stable future direction estimates and identify model-specific error tolerances, ensuring the approximation process stays aligned with the original trajectory without significant deviation.
\begin{figure}[t]
	\centering
	\vspace{-0.1cm}
	\includegraphics[width=1.\linewidth]{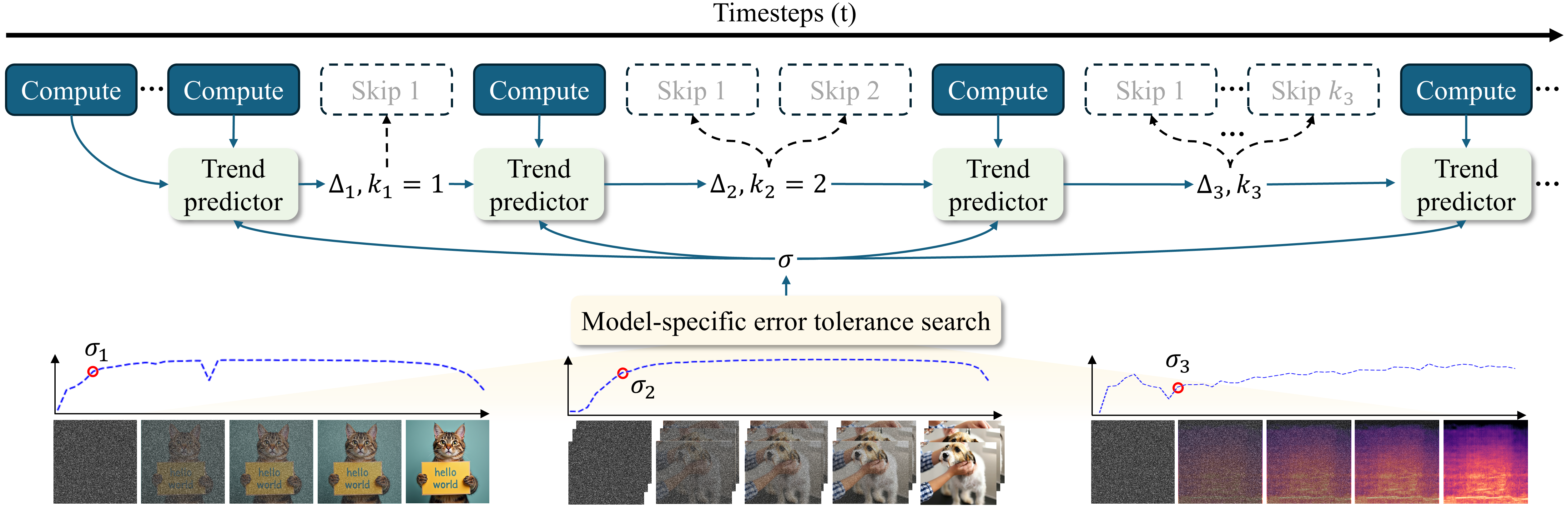}
	\caption{An overview of ETC. ETC leverages all historical model outputs to estimate future trends and dynamically adjusts approximation frequency according to each model’s error tolerance limit.}
	\label{fig:method}
	\vspace{-0.28cm}
\end{figure}

\section{Methodology}
In this section, we provide a detailed description of the architectural components of our proposed framework, ETC. As shown in Figure ~\ref{fig:method}, the overall framework consists of two main modules: 1) Consistent trend predictor for predicting future denoising trends and approximate frequencies; 2) Model-specific error tolerance mechanism for identifying error thresholds across different models. The following subsections detail the design of each module.
\subsection{Preliminary}

\noindent\textbf{Diffusion denoising process}\quad Diffusion models consist of a forward process that progressively corrupts data with Gaussian noise and a reverse process that reconstructs the clean data through iterative denoising. During inference, only the reverse process is executed, starting from the Gaussian noise $x_T\sim \mathcal{N}(0,I)$ and progressively refines $x_T$ into the target output $x_0$ conditioned on input signal $c$ (e.g., text prompts), where $T$ is the predefined number of denoising steps. The refinement follows a generic update rule defined at each timestep $t$ :
\begin{equation}\label{eq:denoise}
	x_{t-1}=f(t-1) \cdot x_{t}-g(t-1) \cdot \epsilon_{\theta}(x_{t}, t, c),\,\,\,\,\,\,\,t=1,\ldots,T, 
\end{equation}
where $\epsilon_{\theta}(x_{t}, t, c)$ is a noise prediction network trained to estimate the noise component in $x_t$ by taking $x_t$, timestep $t$ and an additional condition $c$ as input, while $f(t)$ and $g(t)$ are coefficients determined by the sampler (a.k.a, scheduler) \citep{ho2020denoising,song2020denoising}. Both the noise prediction network $\epsilon_{\theta}$ and sampler coefficients $f(t)$, $g(t)$ directly impact generation quality. Using the same sampler ensures the consistency of $f(t)$ and $g(t)$, so the generation quality is primarily affected by the differences of the noise prediction model $\epsilon_{\theta}$.

\noindent\textbf{Different mathematical definitions of the denoising process}\quad 
The denoising process can be defined through various mathematical frameworks. Stochastic differential equation (SDE)~\citep{song2020score} defines the process as follows:
\begin{equation}\label{eq:sde}
    d x=\left[-\frac{1}{2} \beta(t) x-\nabla_{x_{t}} \log p_{t}\left(x_{t}\right)\right] d t+\sqrt{\beta(t)} d \bar{w},
\end{equation}
where $\nabla_{x_{t}} \log p_{t}\left(x_{t}\right)$ denotes the log-likelihood gradient and $d \bar{w}$ represents the standard Wiener process. Ordinary differential equation (ODE) removes the term $d \bar{w}$ from \autoref{eq:sde}, offering a more stable denoising process that mitigates fluctuations inherent in stochastic processes. Flow matching~\citep{lipman2022flow} proposes a deterministic approach to denoising by learning data flows, mathematically described as follows:
\begin{equation}\label{eq:flow}
    d x=v(x_t,t)d t,
\end{equation}
where $v(x_t,t)$ represents the flow function that governs the evolution of data toward the target distribution over time. Despite their differing formulations, these approaches all ensure the smooth temporal evolution of data features. Therefore, we make the following assumption.

\noindent\textit{Assumption 1}\quad The model output during the denoising process evolves smoothly over time, exhibiting structured and predictable variations across time steps.

\begin{figure}[t]
    \vspace{-0.3cm}
	\centering
	\begin{subfigure}{0.343\textwidth}
		\centering
		\includegraphics[width=\linewidth]{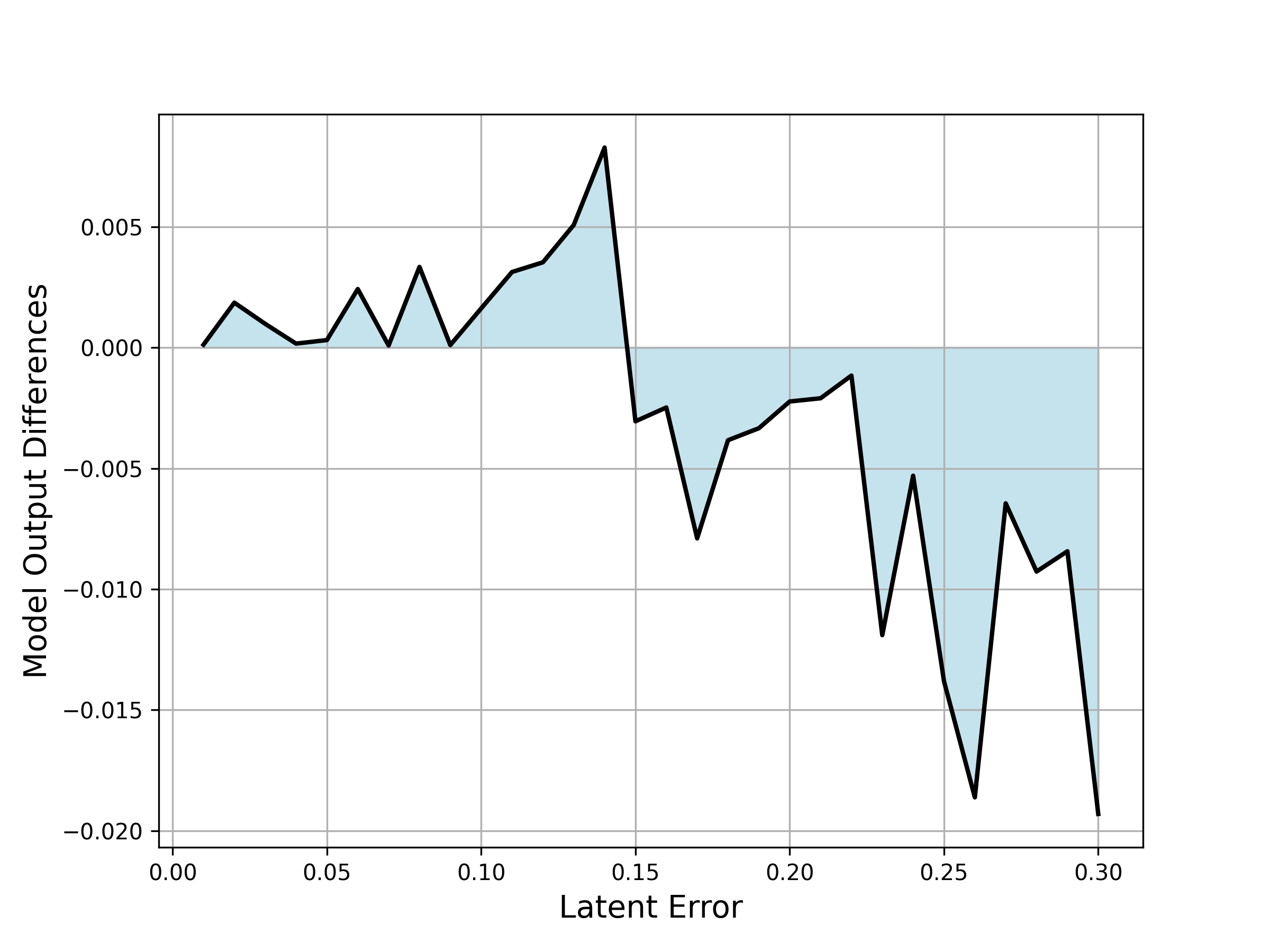}
		\caption{Error correction}
		\label{fig:overpredict}
	\end{subfigure}
	\begin{subfigure}{0.343\textwidth}
		\centering
		\includegraphics[width=\linewidth]{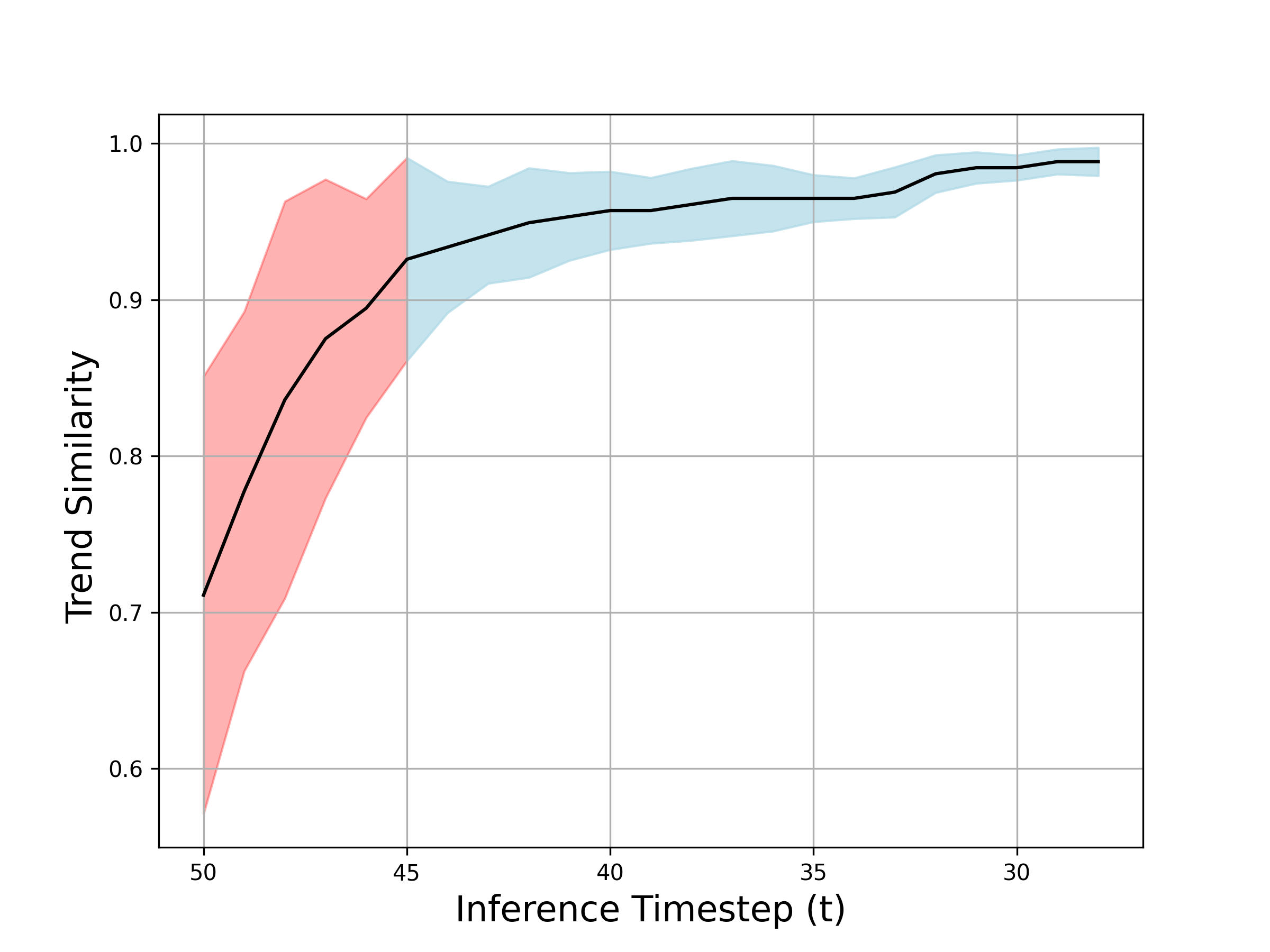}
		\caption{Similarity to historical trends}
		\label{fig:trend-sim}
	\end{subfigure}
	\begin{subfigure}{0.3\textwidth}
		\centering
		\includegraphics[width=\linewidth]{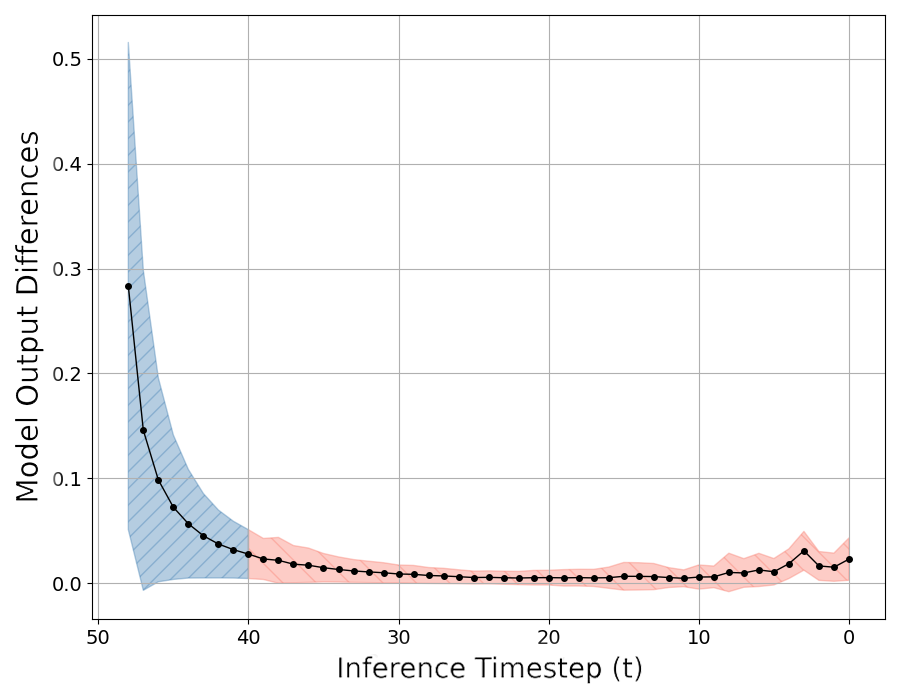}
		\caption{Two stages in denoising}
		\label{fig:stable-stage}
	\end{subfigure}
	\caption{The patterns of the denoising process observed during inference with FLUX on MSCOCO-2017 validation set. Subfigure (a) shows how the model output varies under different latent error. Subfigure (b) illustrates the similarity of current trend to each historical trends. Subfigure (c) depicts the stability of trend changes across different denoising stages.} 
	\vspace{-0.25cm}
	\label{fig:method_stastic}
\end{figure}

\subsection{Consistent trend predictor}
\noindent\textbf{Historical denoising pattern projections}\quad Based on Assumption 1, we can approximate the future denoised trajectory using the recent trend. However, as illustrated in Figure ~\ref{fig:overpredict}, when the approximated trend introduces errors in the latent space, the model output increases compared to the original result in order to correct the cumulative error. Continuing to use this trend for future trajectory approximations leads to fluctuations and further exacerbates trajectory deviation. Therefore, our goal is to approximate trajectories by combining multiple historical trends, thereby minimizing the fluctuations introduced by the error-corrected model output.

Let $d^{t+1}_{t}=\epsilon_{\theta}(x_{t},t,c)-\epsilon_{\theta}(x_{t+1},t+1,c)$ represent the model outputs difference between time steps $t$ and $t+1$. As shown in Figure ~\ref{fig:trend-sim}, recent historical trends more accurately reflect future changes, while even trends from earlier time steps maintain a similarity above 0.7, indicating they still capture some directional information. Therefore, we use a weighted sum of all historical trends, assigning higher weights to more recent trends. However, storing all historical trends incurs considerable computational overhead. To address this, we propose a recursive historical trend weighting method, formulated as follows:
\begin{equation}
    \Delta_{t-2} = (1-\alpha)\Delta_{t-1} + \alpha d^{t}_{t-1}, 0\leq t<T, \Delta_{T-2} = d^{T}_{T-1},
\label{eq:trend_estimate}
\end{equation}
where $\Delta_{t-2}$ represents the estimated trend starting from time step $t-2$, $\alpha$ is the trend adjustment coefficient used to reduce volatility caused by error correction while preserving consistency with the historical denoised trajectory. Only the model outputs and the final approximation value from each round of multi-step approximation are used to calculate the future estimated trends. Furthermore, as shown in Figure ~\ref{fig:trend-sim}, the correlation between the trend in the initial denoising stage and the future trend shows substantial fluctuations (the red area represents the variance). Therefore, we first allow the model to perform n denoising steps to obtain a more stable trend estimate before initiating the approximation. Once the estimated trend is obtained, we can calculate the approximation output $\epsilon^{'}_{\theta}$ using the following formula:
\begin{equation}
    \epsilon^{'}_{\theta}(x_{t-2},t-2,c) = \epsilon_{\theta}(x_{t-1},t-1,c)+\Delta_{t-2}
\label{eq:approximation}
\end{equation}

\noindent\textbf{Dynamic approximation window expansion strategy}\quad To achieve faster acceleration, our goal is to maximize the approximation frequency while maintaining trajectory consistency. As shown in Figure ~\ref{fig:overpredict}, we subtract a fixed value from the latent to represent the approximation error. Within the error correction range, the model can produce larger outputs to compensate for the reduced denoising caused by the error. However, if the error exceeds this range, the model's output begins to diverge from the original result. Based on this observation, we propose an approximation window expansion strategy. If the cumulative error of the previous iteration is below the threshold, the approximation step for the next iteration increases; otherwise, it decreases. The formula is as follows:
\begin{equation}
    k_{t-2} = 
    \begin{cases} 
    k_{t-1}+1, & |d^{t}_{t-1}-\Delta_{t-1}| < \psi, \\
    k_{t-1}-1,        & |d^{t}_{t-1}-\Delta_{t-1}| \geq \psi,
    \end{cases}
\label{eq:k_update}
\end{equation}
where $k$ is the approximation step and $\psi$ is the threshold. To maintain consistency in multi-step approximations, we distribute the estimated trend averagely across each step. This design ensures that the final approximation direction aligns with the estimated trend, preventing the accumulation of estimation errors that could lead to trajectory deviation. The formula for calculating the approximation output at each step is as follows:
\begin{equation}
    \epsilon^{'}_{\theta}(x_{t-1-s,t-1-s,c}) = \epsilon_{\theta}(x_t-1,t-1,c)+\frac{s}{k_{t-2}}\Delta_{t-2}, 1\leq s \leq k_{t-2}.
\label{eq:k_trend}
\end{equation}

\noindent\textbf{Error estimation}\quad The detailed analysis of estimation errors is provided in Appendix ~\ref{sec:error_estimate}. Assuming $\alpha=0.5$, the formulation for the upper bound of cumulative error is as follows:
\begin{equation}
	\small	
	\begin{aligned}
		error_{r} &\leq \|  
		\sum\limits_{m = 0}^{k_r-1}{}((1+ \frac{k_r-m}{2k_r})\sigma_{r-1} + \frac{k_r-m}{2k_r}\cdot d^{T-n-\sum\limits_{l = 1}^{r-2}{(k_l+1)}}_{T-n-\sum\limits_{l = 1}^{r-1}{(k_l+1)}} - d^{T-n-\sum\limits_{j = 1}^{m-1}{(k_j+1)}}_{T-n-\sum\limits_{j = 1}^{m}{(k_j+1)}-k_r+m}
		)\|.
	\end{aligned}
	\label{eq:eq_19}
\end{equation}

\begin{wrapfigure}{r}{0.43\textwidth} 
	\vspace{-0.65cm}	
	\includegraphics[width=0.43\textwidth]{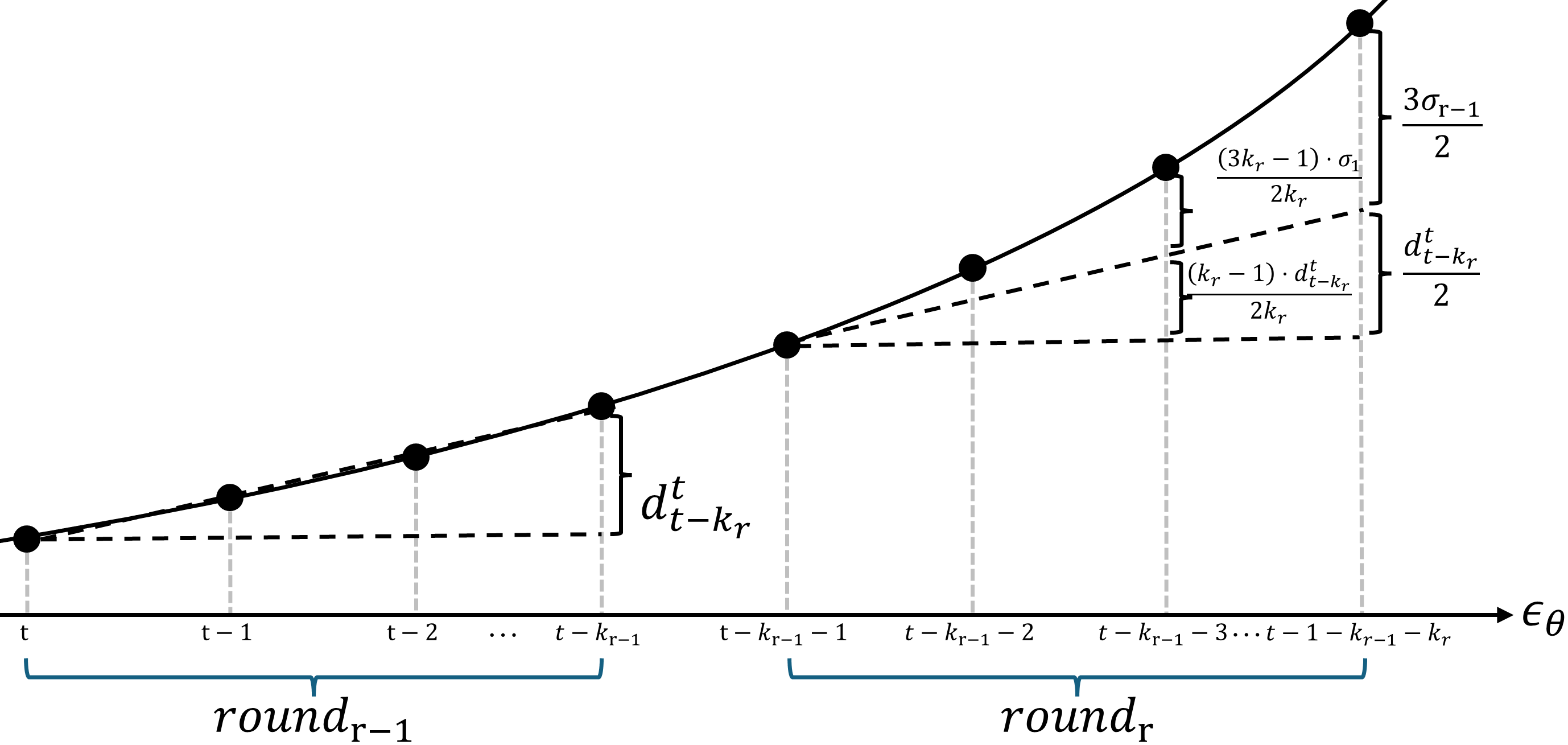} 	
	\caption{Error accumulation at $\alpha=0.5$.}
	\vspace{-0.35cm}
	\label{fig:alpha_error}
\end{wrapfigure} 
Let $t=T-n-\sum\limits_{l = 1}^{r-2}{(k_l+1)}$. As shown in Figure~\ref{fig:alpha_error}, the approximation error depends on the alignment between the approximated trend (a weighted combination of $d$ and $\sigma$) and the future trend. Assuming $\sigma = 0$, the error depends solely on how much the future trend deviates from that of the previous round, ensuring that the denoised trajectory remains at least consistent with the prior step. Consequently, as long as the cumulative error in the early rounds is limited and $\sigma$ remains small, the estimated trend remains controllable and the denoising trajectory maintains consistency.

\subsection{Model-specific error tolerance search mechanism} Previous work \cite{liu2024faster} indicating that cross-attention maps exhibit substantial fluctuations during the semantic planning phase and remain relatively stable throughout the fidelity refinement phase. In alignment with these findings, we observe a similar pattern shown in Figure ~\ref{fig:stable-stage} when analyzing the difference in model outputs between two consecutive steps. We regard the semantic planning phase as fluctuations that occur outside the model's error tolerance range. By identifying the transition point from this fluctuating phase to the stable quality enhancement phase, we can approximate the model's error tolerance limit. Specifically, we introduce the model output difference d from each denoising step as a perturbation to the final latent, while monitoring the similarity between the decoded result and the original generated result. Using trend inflection point analysis tools (e.g., the ruptures package) to analyze the similarity trend, we can determine the model output difference at the transition point between the two phases, which corresponds to our estimated model error tolerance limit. The detailed process is shown in Appendix ~\ref{sec:error_search}.
\begin{table}[t]
\centering
\caption{Quantitative evaluation results. Best performance in \textbf{bold}, and the second best \underline{underlined}. Our method achieves superior efficiency-quality trade-off across diverse tasks and architectures.} 
\label{tab:quanti}
\resizebox{\textwidth}{!}{
\begin{tabular}{c|c c c c|c c c}
\toprule
\textbf{Method} & \multicolumn{4}{c|}{\textbf{Visual Quality}} & \multicolumn{3}{c}{\textbf{Efficiency}} \\
\hline
\hline
\textbf{Text to Image} & \textbf{LPIPS $\downarrow$} & \textbf{SSIM $\uparrow$} & \textbf{PSNR $\uparrow$} & \textbf{CLIP $\uparrow$} & \textbf{FLOPs~(P) $\downarrow$} & \textbf{Speedup $\uparrow$} & \textbf{Latency~(s) $\downarrow$} \\
\hline
\rowcolor[gray]{0.9}
SDXL~($T=50$)  & -  & -  & - & 27.253 & 0.67 & 1$\times$ & 9.63 \\
AdaptiveDiffusion  & 0.172  &0.816  & 25.062 & 27.048 & \underline{0.31} & \underline{$1.98\times$} & \underline{4.87} \\
SADA & \textbf{0.096} &\textbf{0.882}  &\textbf{28.881} & \underline{27.055} & 0.35 & $1.81\times$  & 5.07 \\
Ours ($\sigma=0.0171$) & \underline{0.105} & \underline{0.876} & \underline{28.017} & \textbf{27.103} & \textbf{0.26} & \textbf{2.12$\times$} & \textbf{4.53} \\
\hline
\rowcolor[gray]{0.9}
FLUX~($T=50$)  & -  & -  & - & 26.319 & 3.64 & 1$\times$ & 28.85 \\
TeaCache-fast  & 0.281  &0.753  & 18.845 & 25.924 & \underline{1.45} & \underline{$2.45\times$} & \underline{11.78} \\
SADA & \textbf{0.062} &\underline{0.923}  &\textbf{29.342} & \underline{25.979} & 1.74 & $2.07\times$  &13.94 \\
Ours ($\sigma=0.1269$) & \underline{0.068} & \textbf{0.926} & \underline{29.176} & \textbf{25.983} & \textbf{1.31} & \textbf{2.65$\times$} & \textbf{10.86} \\
\hline
\hline
\textbf{Text to Video} & \textbf{LPIPS $\downarrow$} & \textbf{SSIM $\uparrow$} & \textbf{PSNR $\uparrow$} & \textbf{VBench $\uparrow$} & \textbf{FLOPs~(P) $\downarrow$} & \textbf{Speedup $\uparrow$} & \textbf{Latency~(s) $\downarrow$} \\
\hline
\rowcolor[gray]{0.9}
Open-Sora 1.2~($T=30$)  & -  & -  & - & 77.18\% & 3.15 & 1$\times$ & 45.29 \\
TeaCache-fast  & 0.217  & 0.775  & 20.601 & \underline{76.10\%} & 1.51 & $2.06\times$ & 21.95 \\
MagCache-fast & \underline{0.164} &\underline{0.823}  &\underline{22.583} & 75.84\% & \underline{1.41} & \underline{$2.11\times$}  & \underline{21.45} \\
Ours ($\sigma=0.2101$) & \textbf{0.131} & \textbf{0.848} & \textbf{24.123}  & \textbf{76.44\%} & \textbf{1.32} & \textbf{2.15$\times$} & \textbf{21.07} \\
\hline
\rowcolor[gray]{0.9}
Wan 2.1~($T=50$)  & -  & -  & - & 81.01\% & 76.71 & 1$\times$ & 970.49 \\
TeaCache-fast  & 0.317 & 0.587 & 17.329 & 80.06\% & 38.63 & 1.91$\times$ & 508.31 \\
MagCache-fast & \underline{0.135} & \underline{0.787} & \underline{22.867} & \underline{80.63\%} & \underline{32.49} & \underline{2.38$\times$}  & \underline{407.64} \\
Ours ($\sigma=0.1181$) & \textbf{0.103} & \textbf{0.806} & \textbf{24.078} & \textbf{80.71\%} & \textbf{28.63} & \textbf{2.49$\times$} & \textbf{389.89} \\
\hline
\hline
\textbf{Text to Audio} & \textbf{FAD $\downarrow$} & \textbf{MCD $\downarrow$} & \textbf{KL $\downarrow$} & \textbf{CLAP $\uparrow$} & \textbf{FLOPs~(T) $\downarrow$} & \textbf{Speedup $\uparrow$} & \textbf{Latency~(s) $\downarrow$} \\
\hline
\rowcolor[gray]{0.9}
TangoFlux~($T=50$) & -  & -  & - & 13.286 & 46.86 & 1$\times$ & 5.49 \\
TeaCache-fast  & 0.101  & 3.598  & 0.187 & 13.225 & \underline{16.58} & \underline{$2.27\times$} & \underline{2.42} \\
AdaptiveDiffusion & \underline{0.042} &\underline{3.145}  &\underline{0.181} & \textbf{13.309} & 27.91 & $1.48\times$  & 3.72 \\
Ours ($\sigma=0.0675$) & \textbf{0.026} & \textbf{1.877} & \textbf{0.157} & \underline{13.251} & \textbf{14.83} & \textbf{2.43$\times$} & \textbf{2.26} \\
\bottomrule
\end{tabular}
}
\vspace{-0.2cm}
\end{table}

\section{EXPERIMENTS}

\subsection{Experimental Settings}

\noindent\textbf{Base models and compared methods}\quad
To demonstrate the general effectiveness of our method, we apply our technique to various diffusion models for image, video, and audio generation, including SDXL~\citep{podell2023sdxl} and Flux~\citep{flux2024} for image, Open-Sora 1.2~\citep{opensora} and Wan 2.1~\citep{wan2025wan} for video, and TangoFlux~\citep{hung2024tangoflux} for audio. We compare our method with recent state-of-the-art acceleration approaches, including AdaptiveDiffusion~\citep{ye2024training}, SADA~\citep{jiang2025sada}, TeaCache~\citep{liu2025timestep} and MagCache~\citep{ma2025magcache}. To better assess the efficiency-quality trade-off, we utilize the fast configurations of TeaCache and MagCache for fair comparison.

\noindent\textbf{Evaluation metrics and datasets}\quad 
We evaluate acceleration methods across two key dimensions: computational efficiency and generation quality. For efficiency assessment, we report Floating Point Operations (FLOPs), inference latency and speedup ratios. For quality evaluation, we employ modality-specific metrics to comprehensively assess the fidelity between accelerated and original outputs. For image and video generation, we measure visual similarity using LPIPS~\citep{zhang2018unreasonableeffectivenessdeepfeatures}, PSNR, and SSIM. Additionally, we evaluate text-image alignment using CLIP score~\citep{radford2021learning} for images and employ VBench~\citep{huang2024vbench} for multi-aspect video quality assessment. For audio generation, we measure acoustic fidelity using FAD~\citep{kilgour2018fr}, MCD~\citep{kubichek1993mel}, and KL-divergence of classification probabilities~\citep{copet2023simple,koutini2021efficient}, while assessing text-audio alignment with CLAP score~\citep{wu2023large}. To ensure fair comparison, all experiments are conducted using standardized prompt datasets: MSCOCO-2017~\citep{lin2014microsoft} validation set for image generation, VBench prompts for video generation and AudioCaps~\citep{audiocaps} test set for audio generation.

\textbf{Implementation details}\quad
All experiments are conducted on a same GPU using PyTorch, with FlashAttention~\citep{dao2022flashattention} enabled across all configurations. To efficiently determine the step-skipping threshold, we sample a small set of prompts for each modality. For image and audio models, we partition MSCOCO-2017 and AudioCaps training prompts into 10 length-based bins and sample one prompt per bin. For video models, we sample one prompt from each VBench evaluation dimension. Threshold results are provided in Table~\ref{tab:quanti}. The hyperparameter $n$ is set to 4 for OpenSora and 6 for the remaining models, while $\alpha$ is all set to 0.5.

\begin{figure}[t]
	\centering
	\vspace{-0.1cm}
	\includegraphics[width=1.\linewidth]{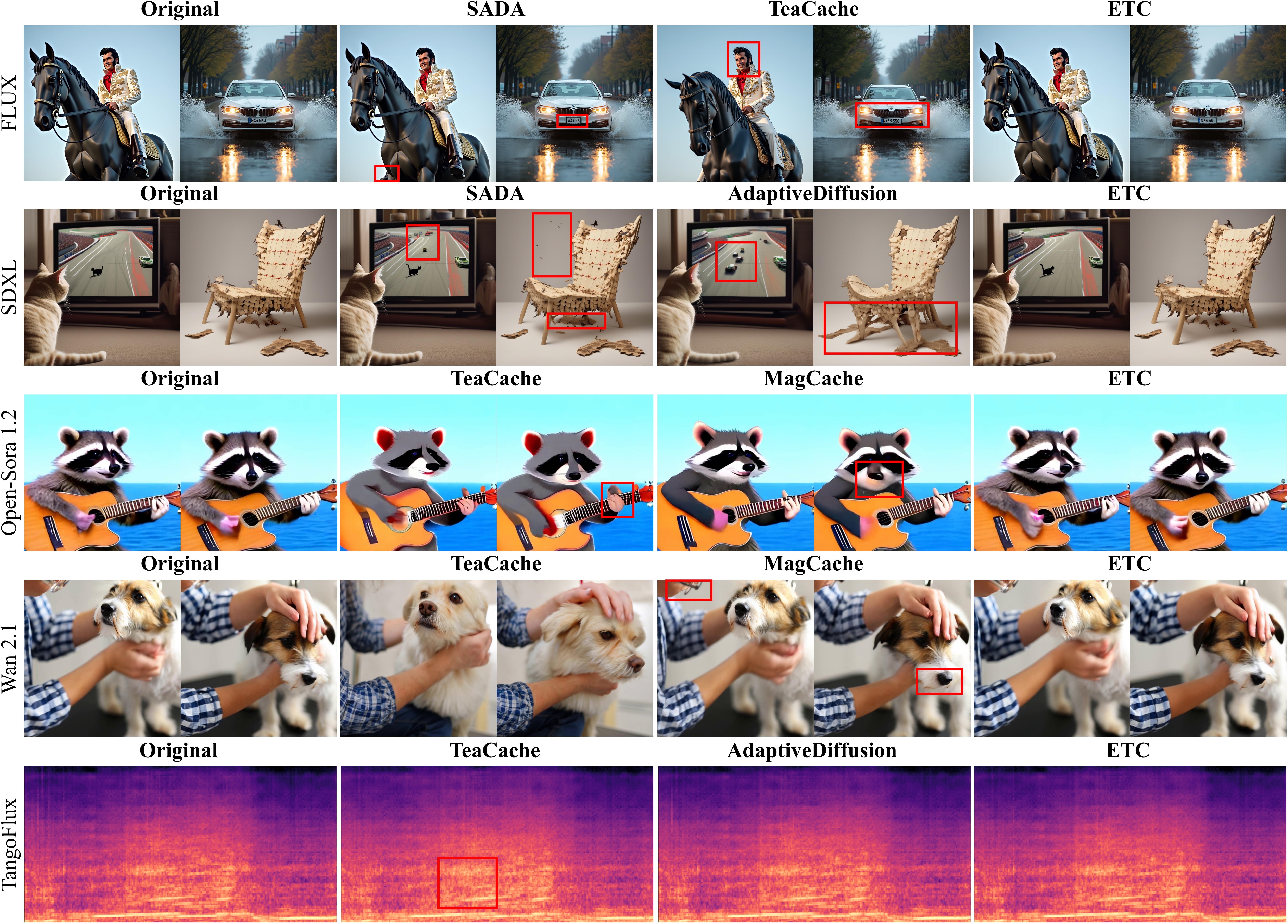}
	\caption{Comparison of visual quality with the competing method. Other methods exhibit issues such as text failure and missing details, whereas ETC achieves the best generation consistency.}
	\label{fig:quali}
	\vspace{-0.28cm}
\end{figure}

\subsection{Main Results}

\noindent\textbf{Quantitative comparison}\quad
Table \ref{tab:quanti} presents quantitative results evaluating both computational efficiency and generation quality. ETC achieves superior acceleration performance across diverse tasks and architectures while maintaining high visual quality. In evaluating the Wan 2.1 text-to-video baseline, our method achieves a 2.5$\times$ speedup while maintaining the highest SSIM of 0.806, outperforming MagCache which attains only 2.38$\times$ acceleration with 5\% lower generation consistency. With the TangoFlux text-to-audio baseline, our approach delivers the highest speedup of 2.43$\times$ with an MCD score 40\% lower than AdaptiveDiff, which achieves only 1.48$\times$ acceleration due to its design prioritizing generation quality. For text-to-image baselines, our method achieves a 2.65$\times$ speedup on Flux with an SSIM of 0.926. Although our SSIM on SDXL is 0.6\% lower than SADA, we achieve 22\% faster inference, demonstrating a superior efficiency-quality trade-off.

\noindent\textbf{Visualization}\quad Figure \ref{fig:quali} compares the results generated by ETC against those by the original model and other baselines. For image generation, TeaCache and AdaptiveDiff exhibit noticeable structural distortions, such as changes in car front designs and chair leg deformations. Although SADA preserves the overall structure, it suffers from detail loss and text generation errors, such as the extra piece of litter near the chair and the altered license plate from "NX4 5KJ" to "AX4 5K". For video generation, TeaCache shows significant content deviation, with differences in dog appearance and petting gestures. MagCache maintains structural alignment but loses fine details such as raccoon fur texture and the distinctive yellow whiskers around the dog's mouth. For audio generation, while the task complexity is relatively lower and most methods preserve similarity to original results, TeaCache still introduces artifacts such as noise distortions highlighted in the red box. In contrast, our method preserves both global consistency and fine-grained detail through various tasks, demonstrating superior generation fidelity.

\begin{table}[t]
	\centering
	\caption{Variation in speed and generation quality under different expansion threshold settings.} 
	\label{tab:sigma}
	\resizebox{\textwidth}{!}{
		\begin{tabular}{c|c|c c c c c c c c c}
			\toprule
			\multirow{4}{*}{FLUX} & $\sigma$ & 0.4429 & 0.2435 & 0.2263 & 0.1921 & \cellcolor{gray!20}0.1269 & 0.1027 & 0.0798 & 0.0597 &0.0392\\ \hhline{~|-|-|-|-|-|-|-|-|-|-|} 
			& Latency (s) $\downarrow$ & 8.64 & 9.06 & 9.61 & 9.62 & \cellcolor{gray!20}10.86 & 11.37 & 11.95 & 13.69 & 14.66 \\ 
			& SSIM $\uparrow$ & 0.823 & 0.851 & 0.887 & 0.893 & \cellcolor{gray!20}0.926 & 0.927 & 0.942 & 0.965 & 0.980 \\
			& CLIP $\uparrow$ & 25.679 & 25.437 & 25.817 & 25.769 & \cellcolor{gray!20}25.983 & 26.281 & 26.169 & 26.323 & 26.494 \\
			\hline
			\hline
			\multirow{4}{*}{Wan 2.1} & $\sigma$ & 0.4225 & 0.2762 & 0.2102 & 0.1523 & \cellcolor{gray!20}0.1181 & 0.0869 & 0.0761 & 0.0678 & 0.0487 \\ \hhline{~|-|-|-|-|-|-|-|-|-|-|} 
			& Latency (s)  $\downarrow$ & 293.76 & 293.78 & 342.91 & 369.97 & \cellcolor{gray!20}389.89  & 496.15 & 505.88 & 536.82 & 538.68 \\ 
			& SSIM $\uparrow$ & 0.637 & 0.637 & 0.708 & 0.762 & \cellcolor{gray!20}0.806  & 0.821 & 0.843 & 0.889 & 0.912 \\
			& VBench $\uparrow$ & 80.61\% & 80.61\% & 80.25\% & 80.56\% & \cellcolor{gray!20}80.71\% & 80.72\% & 80.78\% & 80.95\% & 81.10\% \\
			\bottomrule
		\end{tabular}
	}
\end{table}

\begin{figure}[t]
	\centering
	\begin{subfigure}{0.329\textwidth}
		\centering
		\includegraphics[width=\linewidth]{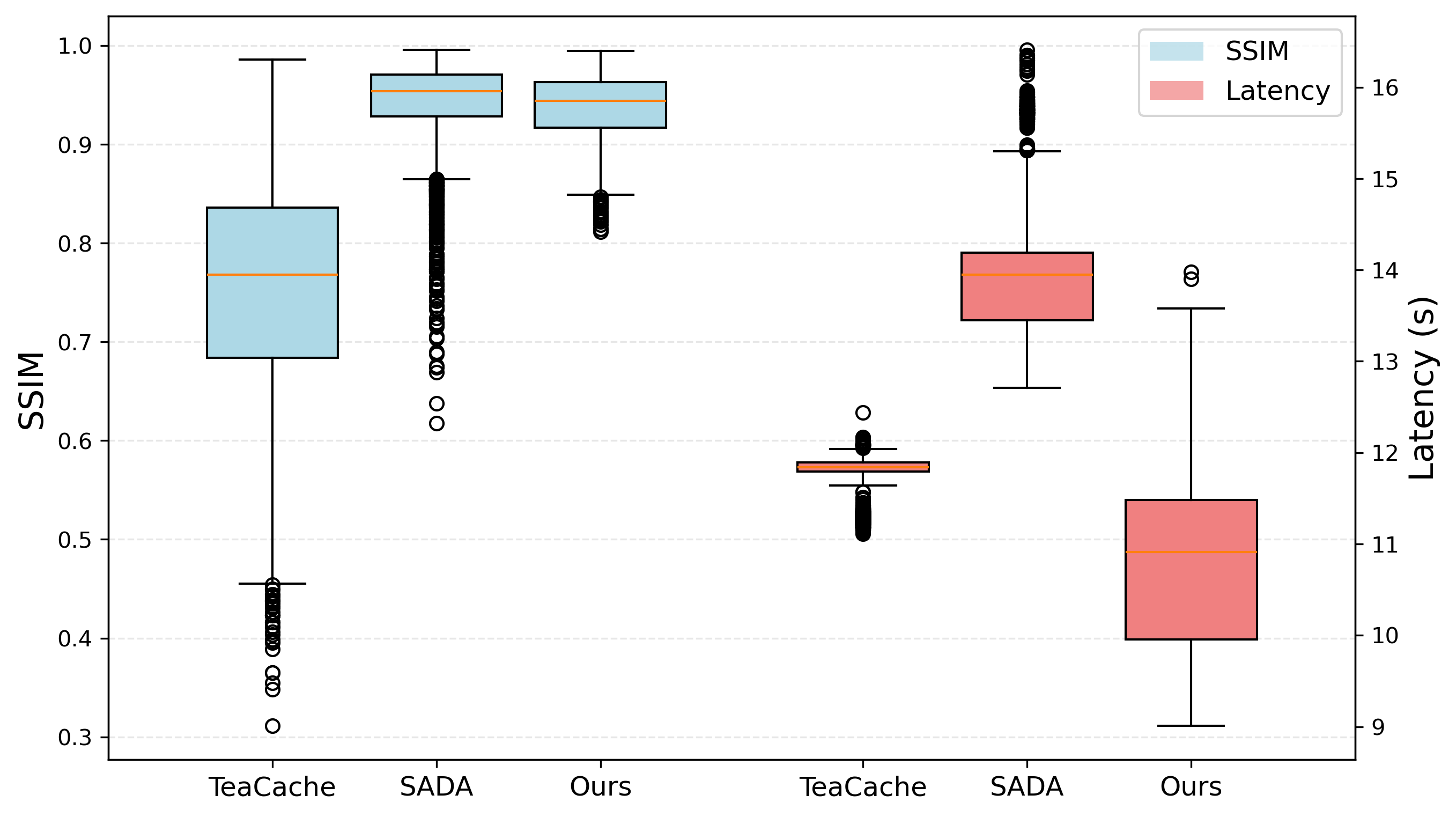}
		\caption{FLUX}
		\label{fig:stastic-flux}
	\end{subfigure}
	\begin{subfigure}{0.329\textwidth}
		\centering
		\includegraphics[width=\linewidth]{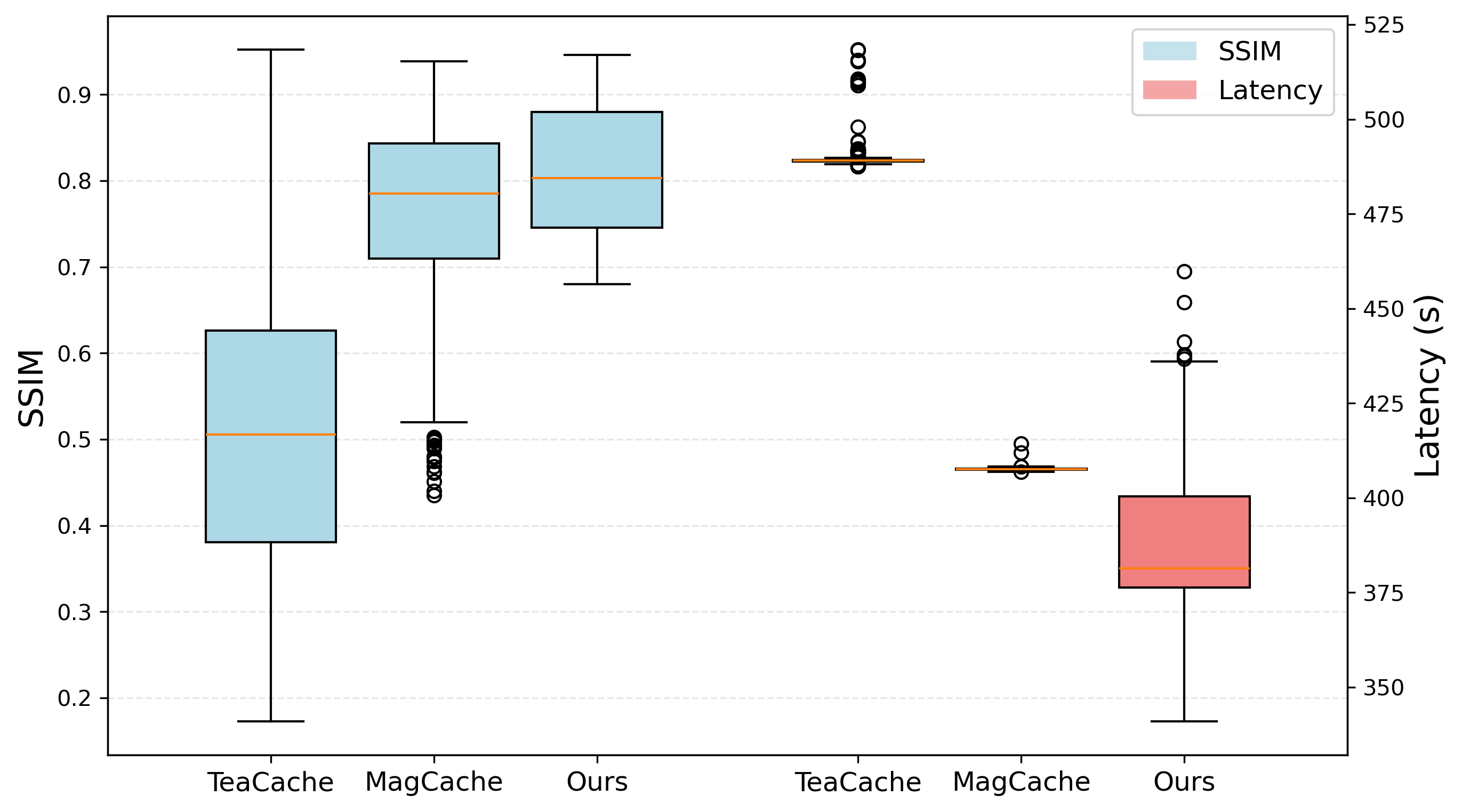}
		\caption{Wan 2.1}
		\label{fig:stastic-wan}
	\end{subfigure}
	\begin{subfigure}{0.329\textwidth}
		\centering
		\includegraphics[width=\linewidth]{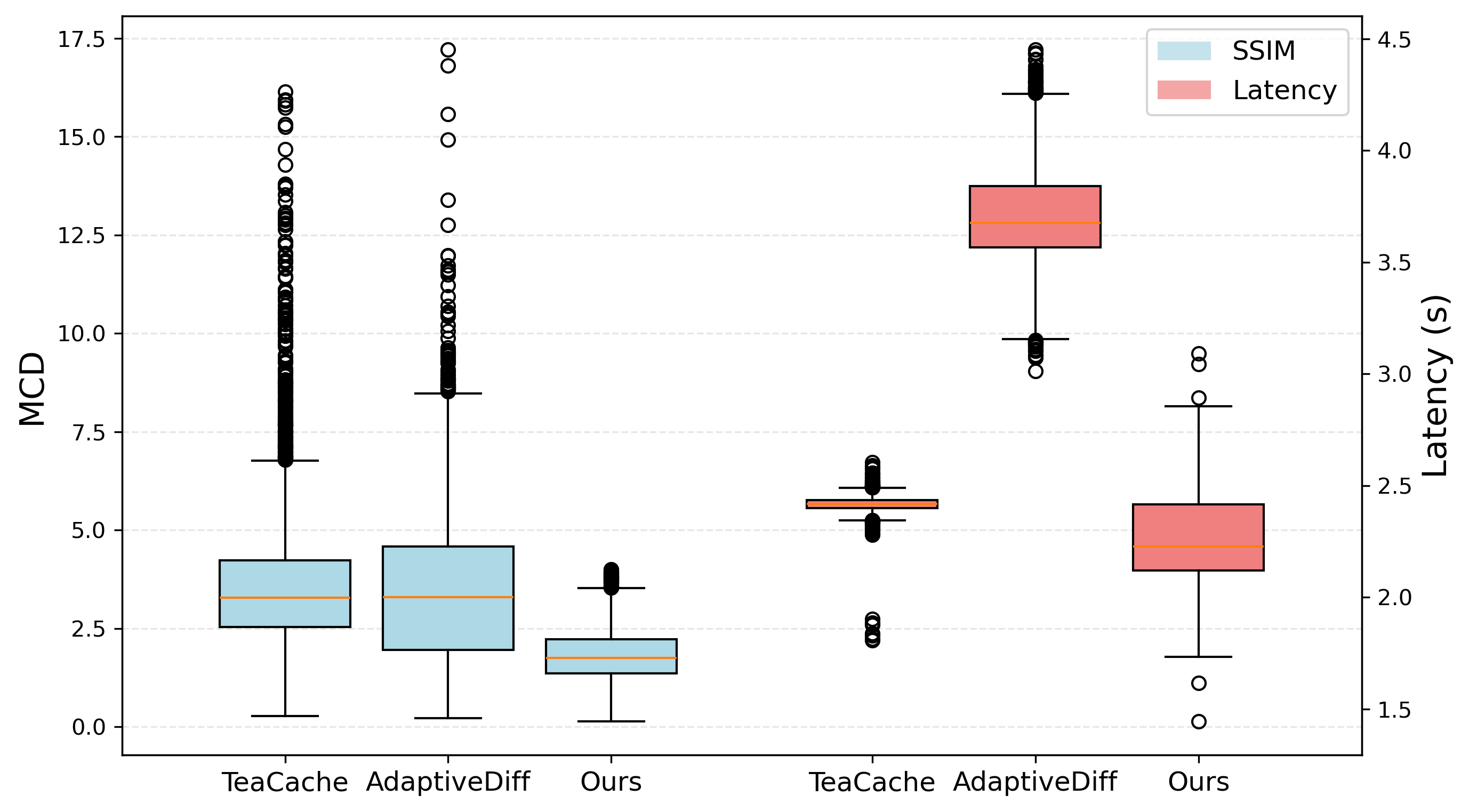}
		\caption{TangoFlux}
		\label{fig:stastic-tangoflux}
	\end{subfigure}
	\caption{Boxplots of SSIM and latency metrics across different tasks.}
	\vspace{-0.2cm}
	\label{fig:stastic}
\end{figure}

\subsection{Ablation Studies}

\noindent\textbf{Statistical robustness}\quad To evaluate the stability of generation performance across diverse text conditions, we analyze the distribution of SSIM and latency metrics using box plots for all three generation tasks. As shown in Figure \ref{fig:stastic}, TeaCache and MagCache exhibit stable acceleration performance with narrow latency distributions across different prompts. However, for Wan 2.1, TeaCache’s SSIM spans a wide range from 0.1 to 0.9, while MagCache shows slight improvement but still ranges from 0.5 to 0.9. In contrast, our method shows moderate latency variance but achieves significantly tighter SSIM distributions of 0.7 to 0.9. While SADA and AdaptiveDiff show similar adaptive behavior with moderate latency variation, both methods suffer from quality instability with numerous outliers compared to our method. These results demonstrate that ETC effectively balances acceleration and stability, achieving more consistent generation quality across diverse conditions.

\begin{wrapfigure}{r}{0.5\textwidth} 
	\vspace{-0.25cm}	
	\includegraphics[width=0.5\textwidth]{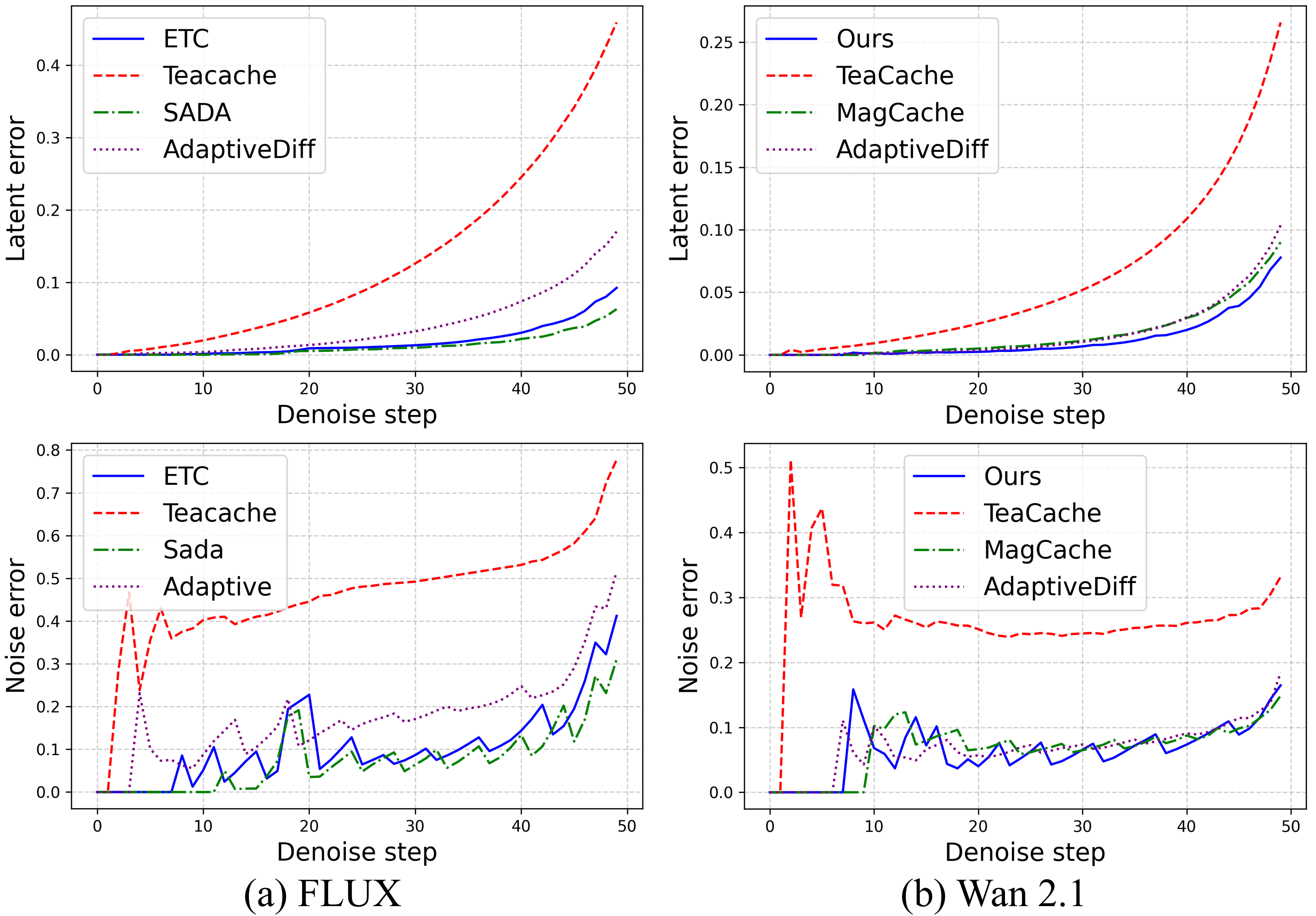} 	
	\caption{Accumulation of denoising errors.}
	\vspace{-0.1cm}
	\label{fig:flux_error}
\end{wrapfigure} 
\noindent\textbf{Effectiveness of error control}\quad To investigate error accumulation from approximate noise, we computed differences between approximated and actual noise at each timestep, along with corresponding latent bias. As shown in Figure \ref{fig:flux_error}, noise error exhibits two distinct patterns: fluctuating increase and steady increase. TeaCache demonstrates steady error growth due to its simplistic approximation using adjacent historical noise differences, introducing significant bias during high-fluctuation initial phases that propagates throughout denoising. Our method follows the fluctuating pattern but accumulates error slower than MagCache and AdaptiveDiff. While SADA exhibits a similar error accumulation pattern to ours, it achieves lower accumulation rates by sacrificing inference speed. In contrsast, our approach maintains high acceleration while effectively controlling error accumulation, demonstrating that our trend approximation strategy preserves denoising trajectory consistency during multiple approximations. 

\noindent\textbf{Effectiveness of threshold selection}\quad As shown in Table \ref{tab:sigma}, we evaluate the acceleration performance of our method under different expansion thresholds. When accelerating different base models, generation quality declines sharply before the optimal threshold but stabilizes afterward. This trend matches the denoising deviation patterns observed in our threshold search, showing that our method effectively captures each model’s tolerance to deviation. Moreover, our method achieves higher SSIM than those obtained during threshold search at high thresholds, indicating that our smoothed trend estimation helps counteract the instability of denoising under high thresholds.

\begin{figure}[t]
	\centering
	\vspace{-0.1cm}
	\includegraphics[width=1.\linewidth]{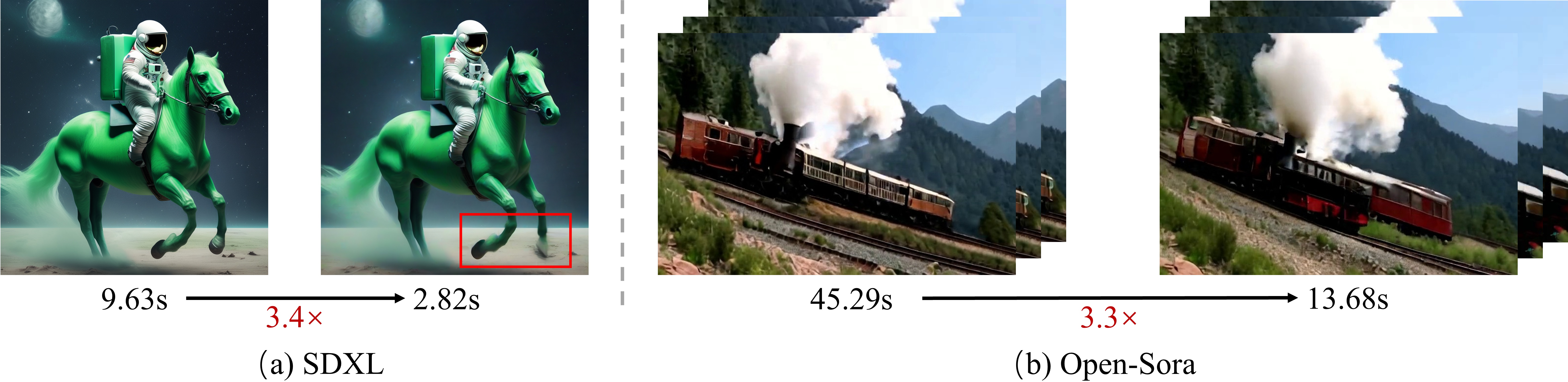}
	\vspace{-0.4cm}
    \caption{Even with less error control for faster acceleration, our method preserves overall structural consistency, with only detail variations.}
	\label{fig:high-speed}
	\vspace{-0.28cm}
\end{figure}

\begin{wraptable}{r}{5.93cm}
	\centering
	\vspace{-0.25cm}
	\caption{Generated results using a fixed error threshold.}
	\resizebox{0.43\textwidth}{!}{
		\begin{tabular}{c|c c c c }
			\toprule
			Model & SDXL & FLUX & Open-Sora 1.2 & Wan 2.1\\
			\hline
			SSIM$\uparrow$ & 0.742 & 0.891 & 0.849 & 0.719\\
			Latency (s)$\downarrow$ & 2.78 & 9.58 & 21.07 & 348.72\\
			Speedup$\uparrow$ & 3.46 & 3.01 & 2.15 & 2.78 \\
			\bottomrule
		\end{tabular}
	}
	\vspace{-0.1cm}
	\label{tab:fix_error}
\end{wraptable}
\noindent\textbf{Fixed error threshold}\quad Table \ref{tab:fix_error} shows the generation results when the error threshold searched from Open-Sora is applied as a fixed threshold across different models. As the gap from the optimal threshold varies, different models exhibit varying degrees of quality degradation. Therefore, employing model-specific error thresholds ensures consistent performance of ETC across different models. Furthermore, as shown in Table~\ref{tab:sigma} and Figure~\ref{fig:high-speed}, when employing larger error thresholds for over 3× acceleration, our method still maintains reasonable similarity (SSIM $>$ 0.65). This demonstrates that even with less error control, our consistent trend estimation preserves the overall structural integrity of the generated results.

\begin{wraptable}{r}{5.93cm}
	\centering
	\vspace{-0.25cm}
	\caption{SSIM results at different $p$ and $\alpha$.}
	\resizebox{0.43\textwidth}{!}{
		\begin{tabular}{c|c c c c c}
			\toprule
			$n$ & 2 & 4 & \cellcolor{gray!20}6 & 7 & 8 \\
			\hline
			FLUX & 0.907 & 0.920 & \cellcolor{gray!20}0.926 & 0.925 & 0.927\\
			Wan 2.1 & 0.722 & 0.786 & \cellcolor{gray!20}0.806 & 0.806 & 0.810\\
			\hline
			$\alpha$ & 0.3 & 0.4 & \cellcolor{gray!20}0.5 & 0.6 & 0.7 \\
			\hline
			FLUX & 0.923 & 0.925 & \cellcolor{gray!20}0.926 & 0.927 & 0.921 \\
			Wan 2.1 & 0.798 & 0.806 & \cellcolor{gray!20}0.806 & 0.803 & 0.791\\
			\bottomrule
		\end{tabular}
	}
	\vspace{-0.1cm}
	\label{tab:pa}
\end{wraptable}
\noindent\textbf{Sensitivity of $n$ and $\alpha$}\quad Table \ref{tab:pa} presents the SSIM variations under different settings of $n$ and $\alpha$. For the parameter $n$, initiating dilation approximation during early denoising stages with high volatility leads to degraded generation consistency. Conversely, larger $n$ provide more stable initial trend estimation but lead to marginal improvements in SSIM. This indicates that permitting a few initial denoising steps helps establish a more stable estimate of the denoising trajectory. Additionally, the results demonstrate minimal sensitivity to different $\alpha$, with SSIM variations remaining within a narrow range across all tested configurations. This robustness indicates that our trend smoothing design effectively maintains the accelerated denoising trajectory aligned with the original sampling path, preventing significant deviation from the intended generation process.

\section{CONCLUSION AND DISCUSSION}
In this work, we proposed ETC, a training-free diffusion acceleration framework that project all historical model outputs into consistent future trends and distributing them across multiple steps within the model’s tolerance limits. Experiments show that ETC preserves generative fidelity while providing substantial speedup. However, a key limitation lies in determining the maximum approximation steps per iteration, where we currently adopt a conservative adjustment strategy based only on the previous round. This restricts the attainable acceleration. A promising direction for future work is to estimate the maximum feasible approximations by evaluating the gap between accumulated errors and model-specific tolerance boundaries.



\bibliography{iclr2026_conference}
\bibliographystyle{iclr2026_conference}

\appendix
\clearpage
\section{Appendix}

\subsection{Error estimation induced by consistent trend predictor}\label{sec:error_estimate}

\subsubsection{A generalized formula for model inference k steps}
Suppose we inference once to obtain model output $\epsilon_{\theta}(x_{t},t,c)$ from the latent $x_{t}$, the denoised formula is as follows:
\begin{equation}
	\small
	x_{t-1} = f(t-1) \cdot x_{t} - g(t-1) \cdot \epsilon_{\theta}(x_{t},t,c)
\label{eq:eq_1}
\end{equation}
Similarly, we can obtain the following formula after the second denoising and substituting ~\autoref{eq:eq_1}:
\begin{equation}
	\small
	\begin{aligned}
		x_{t-2} &= f(t-2) \cdot x_{t-2} - g(t-2) \cdot \epsilon_\theta(x_{t-1}, t-1, c)
		\\
		&= f(t-2) \cdot f(t-1) \cdot x_{t} - f(t-2) \cdot g(t-1) \cdot \epsilon_\theta(x_{t}, t, c) - g(t-2) \cdot \epsilon_\theta(x_{t-1}, t-1, c).
	\end{aligned}
\label{eq:eq_2}
\end{equation}

By analogy, we can obtain the following results after sampling k times using the model output:
\begin{equation}
	\small
	\begin{aligned}
		x_{t-k} &= \prod\limits_{{\rm{j}} = t - k}^{t - 1} {f(j) \cdot {x_{t}}} - g(t-k) \cdot \epsilon_\theta(x_{t-k+1}, t-k+1, c) 
		\\
		&\,\,\,\,\,\, - \sum\limits_{m = 0}^{k - 2}{((\prod\limits_{j = t-k}^{t-k+m}{f(j)})\cdot g(t-k+m+1) \cdot \epsilon_\theta(x_{t-k+m+2}, t-k+m+2, c))}.
	\end{aligned}
\label{eq:eq_3}
\end{equation}

\subsubsection{Cumulative error caused by using estimated trends for the first time}\label{sec:first_apply_kstep} 
\begin{figure}[h]
	\centering
	\includegraphics[width=0.9\linewidth]{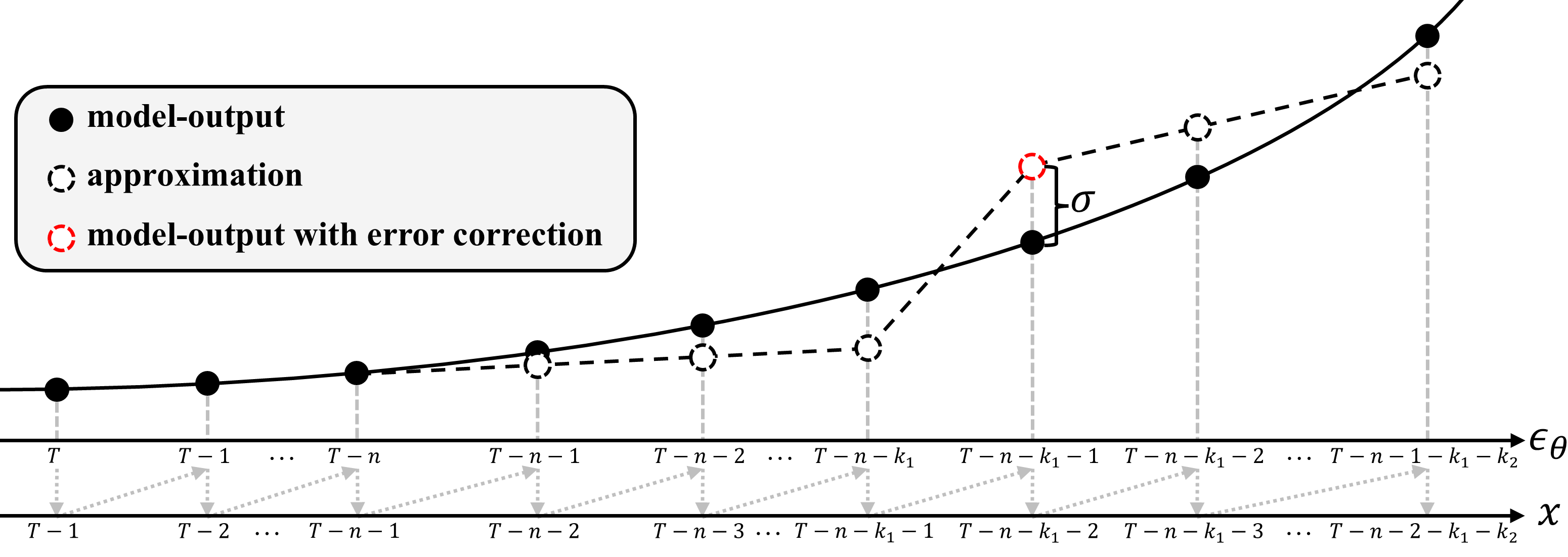}
	\caption{Error accumulation for $k$ times approximation used after model inference $n$ times.}
	\label{fig:appendix_1}
	\vspace{-0.1cm}
\end{figure}

\noindent\textbf{Assumptions}\quad As shown in Figure ~\ref{fig:appendix_1}, we make the assumptions that:

(1) The timestep of the denoising process decreases from T to 0;

(2) Let $ d^{t_1}_{t_2} = \epsilon_\theta(x_{t_2}, t_2, c) - \epsilon_\theta(x_{t_1}, t_1, c) $;

(3) Estimated trend is utilized for denoising after performing model inference $n$ times.

\noindent\textbf{Initial approximate trend}\quad When the model performs inference twice first and then the outputs are used to calculate the estimated trend, the formula is as follows:
\begin{equation}
	\small	
	\Delta_{T-2} = \epsilon_\theta(x_{T-1}, T-1, c) - \epsilon_\theta(x_{T}, T, c) = d^{T}_{T-1}.
\label{eq:eq_4}
\end{equation}

The second estimated trend formulation is as follows:
\begin{equation}
	\small	
	\Delta_{T-3} = (1-\alpha)\cdot\Delta_{T-2} + \alpha \cdot d^{T-1}_{T-2} = (1-\alpha)\cdot d^{T}_{T-1} + \alpha\cdot d^{T-1}_{T-2}.
\label{eq:eq_5}
\end{equation}

The third estimated trend formulation is as follows:
\begin{equation}
	\small	
	\begin{aligned}
		\Delta_{T-4} &= (1-\alpha)\cdot\Delta_{T-3} + \alpha \cdot d^{T-2}_{T-3} = (1-\alpha)\cdot ((1-\alpha)\cdot d^{T}_{T-1} + \alpha\cdot d^{T-1}_{T-2}) + \alpha\cdot d^{T-2}_{T-3}
		\\
		&= (1-\alpha)^2\cdot d^{T}_{T-1} +\alpha\cdot(1-\alpha)\cdot d^{T-1}_{T-2}+\alpha\cdot d^{T-2}_{T-3}
	\end{aligned}.
\label{eq:eq_6}
\end{equation}

By analogy, we can obtain the following estimated trend for the $n^{th}$ iteration:
\begin{equation}
	\small	
	\Delta_{T-n-1} = (1-\alpha)^{n-1}\cdot d^{T}_{T-1} + \alpha \sum\limits_{m = 0}^{n-2}{(1-\alpha)^{m}\cdot d^{T-n+m+1}_{T-n+m}}.
\label{eq:eq_7}
\end{equation}

\noindent\textbf{Cumulative error}\quad To improve readability, we set $\epsilon_\theta(x_{t})=\epsilon_\theta(x_{t},t,c)$. Starting from $x_{T-n-1}$, the formulation for $x_{T-n-1-k_1}$ based on the estimated trend and ~\autoref{eq:eq_3} is as follows:
\begin{equation}
	\small	
	\begin{aligned}
		x^{'}_{T-n-1-k_1} &= \prod\limits_{{\rm{j}} = T-n-1-k_1}^{T-n-2} {f(j) \cdot {x_{T-n-1}}} - g(T-n-1-k_1) \cdot (\epsilon_\theta(x_{T-n})+\Delta_{T-n-1}) 
		\\
		&\,\,\,\,\,\, - \sum\limits_{m = 0}^{k_1 - 2}{((\prod\limits_{j = T-n-1-k_1}^{T-n-1-k_1+m}{f(j)})\cdot g(T-n-k_1+m) \cdot (\epsilon_\theta(x_{T-n})+\frac{k_1-m-1}{k_1}\Delta_{T-n-1}))}.
	\end{aligned}
\label{eq:eq_8}
\end{equation}

The cumulative error is as follows:
\begin{equation}
	\small	
	\begin{aligned}
		error &= \|x_{T-n-1-k_1} - x^{'}_{T-n-1-k_1}\|
		\\
		&= \|g(T-n-1-k_1)\cdot(\Delta_{T-n-1}-d^{T-n}_{T-n-k_1})
		\\		
		&+\sum\limits_{m = 0}^{k_1 - 2}{}((\prod\limits_{j = T-n-1-k_1}^{T-n-1-k_1+m}{f(j)})\cdot g(T-n-k_1+m) \cdot (\frac{k_1-m-1}{k_1}\Delta_{T-n-1} - d^{T-n}_{T-n-k_1+m+1}))\|.
	\end{aligned}
\label{eq:eq_9}
\end{equation}

Since $f\leq 1$ and $g\leq 1$, we can get the formulation below: 
\begin{equation}
	\small	
	\begin{aligned}
		error &\leq \|(\Delta_{T-n-1}-d^{T-n}_{T-n-k_1})+(\frac{k_1-m-1}{k_1}\Delta_{T-n-1} - d^{T-n}_{T-n-k_1+m+1})\|
		\\
		&=\|\sum\limits_{m = 0}^{k_1-1}{(\frac{k_1-m}{k_1}\Delta_{T-n-1} - d^{T-n}_{T-n-k_1+m})}\|.
	\end{aligned}
\label{eq:eq_10}
\end{equation}

Substituting ~\autoref{eq:eq_7} into ~\autoref{eq:eq_10}, we obtain the following upper bound on the error:
\begin{equation}
	\small	
	\begin{aligned}
		error &\leq \|\sum\limits_{m = 0}^{k_1-1}{(\frac{(k_1-m)(1-\alpha)^{n-1}}{k_1}\cdot d^{T}_{T-1} + \frac{\alpha(k_1-m)}{k_1}\sum\limits_{j = 0}^{n-2}{(1-\alpha)^{j}}\cdot d^{T-n+j+1}_{T-n+j} -
d^{T-n}_{T-n-k_1+m})}\|.
	\end{aligned}
\label{eq:eq_11}
\end{equation}

\subsubsection{Error accumulation in the next round}\label{sec:second_apply_kstep} 
\noindent\textbf{Assumptions}\quad As shown in Figure ~\ref{fig:appendix_1}, we make the assumptions that:

(1) To correct the cumulative error, the model output $\epsilon^{*}_\theta(x_{T-n-k_1-1}) = \epsilon_\theta(x_{T-n-k_1-1})+\sigma_1$. After a single model inference, the obtained $x_{T-n-k_1-2}$ exhibits no error. 

\noindent\textbf{Updated approximate trend}\quad After correcting the cumulative error using a single model inference, we update the formula for estimated trend as follows:
\begin{equation}
	\small	
	\begin{aligned}
		\Delta_{T-n-2-k_1} &= (1-\alpha)\cdot\Delta_{T-n-1} + \alpha\cdot((\epsilon_\theta(x_{T-n-k_1-1})+\sigma_1) - (\epsilon_\theta(x_{T-n})+\Delta_{T-n-1}))
		\\
		&=(1-2\alpha)\cdot\Delta_{T-n-1}+\alpha\cdot(d^{T-n}_{T-n-k_1-1}+\sigma_1).
	\end{aligned}
\label{eq:eq_12}
\end{equation}

\noindent\textbf{Cumulative error}\quad To improve readability, we omit the arguments of $f$ and $g$. Starting from $x_{T-n-2-k_1}$, the formulation for $x_{T-n-2-k_1-k_2}$ based on the estimated trend and ~\autoref{eq:eq_3} is:
\begin{equation}
	\small
	\begin{aligned}
		x^{'}_{T-n-2-k_1-k_2} &= \prod\limits_{}^{} {f \cdot {x_{T-n-2-k_1}}} - g \cdot (\epsilon_\theta(x_{T-n-1-k_1})+\sigma_1 + \Delta_{T-n-2-k_1}) 
		\\
		&\,\,\,\,\,\, - \sum\limits_{m = 0}^{k_2 - 2}{((\prod\limits_{}^{}{f})\cdot g \cdot (\epsilon_\theta(x_{T-n-1-k_1})+\sigma_1+\frac{k_2-m-1}{k_2}\Delta_{T-n-2-k_1}))}.
	\end{aligned}
\label{eq:eq_13}
\end{equation}

The cumulative error is as follows:
\begin{equation}
	\small
	\begin{aligned}
		error &= \|x_{T-n-2-k_1-k_2} - x^{'}_{T-n-2-k_1-k_2}\| = \|g\cdot(\sigma_1+\Delta_{T-n-2-k_1}-d^{T-n-1-k_1}_{T-n-1-k_1-k_2})
		\\		
		&\,\,\,\,\,\, +\sum\limits_{m = 0}^{k_2 - 2}{((\prod\limits_{}^{}{f})\cdot g \cdot (\sigma_1+\frac{k_2-m-1}{k_2}\Delta_{T-n-2-k_1} - d^{T-n-1-k_1}_{T-n-k_1-k_2+m}))}\|.
	\end{aligned}
\label{eq:eq_14}
\end{equation}

Since $f\leq 1$ and $g\leq 1$, we can get the formulation below: 
\begin{equation}
	\small	
	\begin{aligned}
		error &\leq \|(\sigma_1+\Delta_{T-n-2-k_1}-d^{T-n-1-k_1}_{T-n-1-k_1-k_2})+(\sigma_1+\frac{k_2-m-1}{k_2}\Delta_{T-n-2-k_1} -d^{T-n-1-k_1}_{T-n-k_1-k_2+m})\|
		\\
		&=\|\sum\limits_{m = 0}^{k_2-1}{(\sigma_1+\frac{k_2-m}{k_2}\Delta_{T-n-2-k_1} - d^{T-n-1-k_1}_{T-n-1-k_1-k_2+m} )}\|.
	\end{aligned}
\label{eq:eq_15}
\end{equation}

Substituting ~\autoref{eq:eq_12} into ~\autoref{eq:eq_15}, we obtain the following upper bound on the error:
\begin{equation}
	\small	
	\begin{aligned}
		error &\leq \| \sum\limits_{m = 0}^{k_2-1}{}(
			\frac{(k_2-m)(1-2\alpha)(1-\alpha)^{n-1}}{k_2}\cdot d^{T}_{T-1} + \frac{\alpha(k_2-m)(1-2\alpha)}{k_2}\sum\limits_{j = 0}^{n-2}{(1-\alpha)^{j}\cdot d^{T-n+j+1}_{T-n+j}}
		\\		
		&\,\,\,\,\,\, +\frac{\alpha(k_2-m)}{k_2}\cdot d^{T-n}_{T-n-k_1-1} + \frac{k_2+\alpha(k_2-m)}{k_2}\cdot \sigma_1 - d^{T-n-1-k_1}_{T-n-1-k_-k_2+m}
		)\|.
	\end{aligned}
\label{eq:eq_16}
\end{equation}

\subsubsection{Accumulation of errors throughout the process}\label{sec:total_error} 
\begin{figure}[h]
	\centering
	\includegraphics[width=0.95\linewidth]{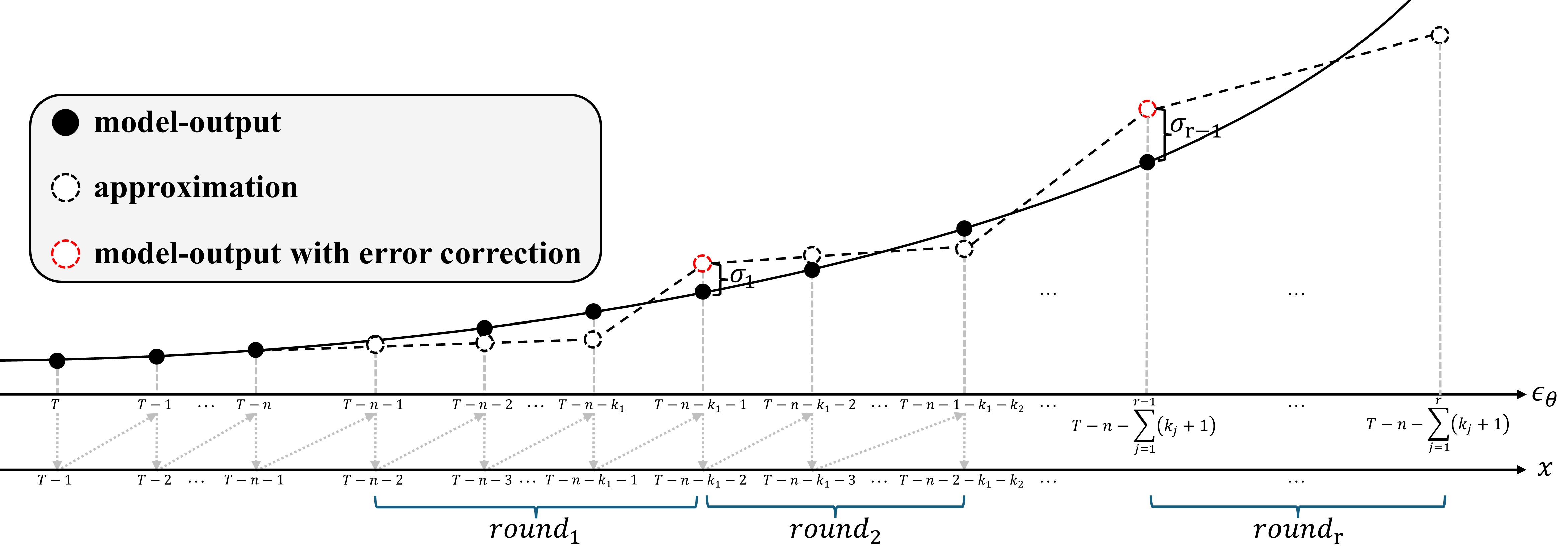}
	\caption{Error accumulation throughout the denoising process.}
	\label{fig:appendix_2}
\end{figure}
\noindent\textbf{Assumptions}\quad As shown in Figure ~\ref{fig:appendix_2}, we make the assumptions that:

(1) A total of $r$ rounds of approximations were performed, with $k_i$ approximations per round, and the difference between the outputs obtained using model inference and the model outputs under no error for each round is $\sigma_{i-1}$, where $i=1,...,r$.

\noindent\textbf{Approximate trend at each round}\quad By analogy, we can obtain the following formulation for the estimated trend used in the $r^{th}$ round:
\begin{equation}
	\small	
	\Delta_{T-n-1-\sum\limits_{m = 1}^{r-1}{(k_{m}+1)}} = (1-2\alpha)^{r-1}\Delta_{T-n-1} + \alpha \sum\limits_{m = 1}^{r-1}{((1-2\alpha)^{r-m-1}(d^{T-n-\sum\limits_{j = 1}^{m-1}{(k_{j}+1)}}_{T-n-\sum\limits_{j=1}^{m}{k_j+1}}+\sigma_{m}))}
\label{eq:eq_17}
\end{equation}

\noindent\textbf{Cumulative error}\quad By analogy, we can obtain the following formulation for the cumulative error in the $r^{th}$ round:
\begin{equation}
	\small	
	\begin{aligned}
	error_{r} &\leq \|  \sum\limits_{m = 0}^{k_r-1}{(\sigma_{r-1}+ \frac{k_r-m}{k_r}\Delta_{T-n-1-\sum\limits_{j = 1}^{r-1}{(k_j+1)}}-d^{T-n-\sum\limits_{j = 1}^{r-1}{(k_j+1)}}_{T-n-\sum\limits_{j = 1}^{r-1}{(k_j+1)}-k_r+m})} \|
	\\
	&= \| \sum\limits_{m = 0}^{k_r-1}{}( \sigma_{r-1} + \frac{(k_r-m)(1-2\alpha)^{r-1}}{k_r}\Delta_{T-n-1} 
	\\
	&\,\,\,\,\,\, + \frac{\alpha(k_r-m)}{k_r}\sum\limits_{j = 1}^{r-1}{(
		(1-2\alpha)^{r-j-1}(d^{T-n-\sum\limits_{l = 1}^{j-1}{(k_l+1)}}_{T-n-\sum\limits_{l = 1}^{j}{(k_l+1)}}+\sigma_j)
	)}
	-d^{T-n-\sum\limits_{j = 1}^{r-1}{(k_j+1)}}_{T-n-\sum\limits_{j = 1}^{r-1}{(k_j+1)}-k_r+m}
	) \|
	\end{aligned}
\label{eq:eq_18}
\end{equation}

Substituting ~\autoref{eq:eq_7} into ~\autoref{eq:eq_18}, we obtain the following upper bound on the error:
\begin{equation}
	\small	
	\begin{aligned}
		error_{r} &\leq \|  
		\sum\limits_{m = 0}^{k_r-1}{}(( \sigma_{r-1} + \frac{\alpha(k_r-m)}{k_r} \sum\limits_{j = 1}^{r-1}{(1-2\alpha)^{r-j-1}\cdot \sigma_{j}})
		\\
		&+( \frac{(k_r-m)(1-2\alpha)^{r-1}(1-\alpha)^{n-1}}{k_r}\cdot d^{T}_{T-1} + \frac{\alpha(k_r-m)(1-2\alpha)^{r-1}}{k_r} \sum\limits_{j = 0}^{n-2}{(1-\alpha)^j\cdot d^{T-n+j+1}_{T-n+j}} 
		\\
		&+ \frac{\alpha(k_r-m)}{k_r}\sum\limits_{j = 1}^{r-1}{(1-2\alpha)^{r-j-1}\cdot d^{T-n-\sum\limits_{l = 1}^{j-1}{(k_l+1)}}_{T-n-\sum\limits_{l = 1}^{j}{(k_l+1)}}} - d^{T-n-\sum\limits_{j = 1}^{m-1}{(k_j+1)}}_{T-n-\sum\limits_{j = 1}^{m}{(k_j+1)}-k_r+m}
		)\|
	\end{aligned}
	\label{eq:eq_19}
\end{equation}

As can be seen, we approximate the future trend at each step by a weighted combination of the historical trend and the error-correction. 


\subsection{Algorithm of consistent trend predictor}\label{sec:alg_trend}
\begin{algorithm}[H]
	\caption{Consistent Trend Predictor.}
	\label{alg:k-step}
	\renewcommand{\algorithmicensure}{\textbf{Input:}}
	\begin{algorithmic}[1]
		\Ensure{Diffusion Model $\epsilon_\theta$, Sampling Scheduler $\phi$, Decoder $D$, Sample Step $T$, Conditional Embedding $c$, Pre-Inference step $n$, Smoothing Factor $\alpha$, Error Threshold $\sigma$;}
		\State Initialize Random Noise $x$, Future Trend $\Delta = $ None, Approximation Step $k$ = 0, Previous Model Output $P=$None.
		\For{$t=T$ to $T - n$}
        \If{$t=T-1$}
        \State $\Delta=\epsilon_\theta(x,t,c) - P$;
        \EndIf
        \If{$t<T-1$}
        \State $\Delta = (1-\alpha)\Delta+\alpha(\epsilon_\theta(x,t,c) - P)$;
        \EndIf
		\State $P = \epsilon_\theta(x,t,c)$);
		\State Compute $\phi(x,\epsilon_\theta(x,t,c))$ by Eq. (\ref{eq:denoise});
		\EndFor
		\While{$t>1$}
		\For{$j = 1$ to $k$}
		\State $\epsilon_\theta^{'}(x, t-j,c) \leftarrow P+ \frac{\Delta}{k}$;
		\State $P=\epsilon^{'}_\theta(x, t-j,c)$);
		\State Compute $\phi(x,\epsilon^{'}_\theta(x, t-j,c))$ by Eq. (\ref{eq:denoise});
		\EndFor
		\State $t \leftarrow t-k-1$;
        \If{$\epsilon_\theta(x, t, c)-P-\Delta \geq \sigma$}
        \If{$k>0$}
        \State $k= k-1$;
        \EndIf
        \EndIf
        \If{$\epsilon_\theta(x, t, c)-P-\Delta < \sigma$}
        \State $k = k+1$;
        \EndIf
        \State $\Delta = (1-\alpha)\Delta+\alpha(\epsilon_\theta(x, t, c)-P)$;
		\State $P=\epsilon_\theta(x, t, c)$);
		\State Compute $\phi(x,\epsilon_\theta(x, t,c))$ by Eq. (\ref{eq:denoise});
		\EndWhile\\
        Compute $\phi(x,\epsilon_\theta(x, t-1,c))$ by Eq. (\ref{eq:denoise});\\
		\Return{$D(x)$.}
	\end{algorithmic}
\end{algorithm}

\subsection{Algorithm of consistent trend predictor}\label{sec:error_search}
\begin{algorithm}[H]
	\caption{Model-Specific Error Tolerance Search Mechanism.}
	\label{alg:k-step}
	\renewcommand{\algorithmicensure}{\textbf{Input:}}
	\begin{algorithmic}[1]
		\Ensure{Diffusion Model $\epsilon_\theta$, Sampling Scheduler $\phi$, Decoder $D$, Sample Step $T$, Conditional Embedding  Set $S_{c}$, Trend Inflection Point Analysis Model $M$, Similarity Metrics $Sim$;}
		\State Initialize Similarity List S=[0]*($T-1$).
		\For{$c$ in $S_{c}$}
       \State Initialize Random Noise $x$, Previous Output $P$ = None, Model Output Differences List $L$=[];
      \For {$t=T$ to 0}
        \If{$t<T$}
        \State $L$.append($(\epsilon_\theta(x, t,c)-P).abs().mean()$);
        \EndIf
        \State $P = \epsilon_\theta(x, t,c)$;
        \State Compute $\phi(x,\epsilon_\theta(x, t,c))$ by Eq. (\ref{eq:denoise});
        \EndFor
        \For{ $i=0$ to $T-1$}
        \State Initialize Random Noise $x^{'}$;
        \State $x^{'} = x^{'}/x^{'}.mean()*L[i]$;
        \State S[$i$] = S[$i$]+$Sim(D(x+x'),D(x))$;
        \EndFor
        \EndFor
        \State S = S/($T$-1);
        \State $\sigma$ = $M$(S);
        \Return{$\sigma$}
	\end{algorithmic}
\end{algorithm}




\end{document}